\documentclass{article}

\usepackage{authblk}  
\usepackage{times}
\usepackage{amsmath}
\usepackage{amsthm}
\usepackage{amssymb}
\usepackage{amscd}
\usepackage{mathrsfs}
\usepackage{latexsym}
\usepackage{dcolumn}
\usepackage{endnotes}
\usepackage{tabularx}
\usepackage{color}
\usepackage{colortbl}
\usepackage{bm}
\usepackage{chemarrow}
\usepackage{CJK}
\usepackage{BOONDOX-cal}
\usepackage[colorlinks,linkcolor=blue,anchorcolor=blue,citecolor=blue]{hyperref}
\usepackage{diagbox}
\usepackage{multirow}
\usepackage{float}
\usepackage{graphicx}
\usepackage{graphics}
\usepackage{epstopdf}
\usepackage{subcaption} 
\usepackage{slashbox}
\usepackage{enumerate}
\usepackage{enumitem}
\usepackage{bbding}
\usepackage{pifont}
\usepackage[numbers,sort&compress]{natbib}
\usepackage[utf8]{inputenc}
\usepackage{booktabs}
\usepackage{setspace}
\usepackage{rotating}
\usepackage[font=small, labelfont=bf, labelsep=none]{caption}
\usepackage[ruled,linesnumbered]{algorithm2e}
\usepackage{caption}
\usepackage{setspace}
\usepackage[table]{xcolor}
\usepackage{tabularx}
\usepackage{makecell}
\usepackage{rotating}
\usepackage{array} 
\usepackage{longtable}
\usepackage{arydshln}

\newtheorem{theorem}{Theorem}[section]

\newtheorem{proposition}{Proposition}[section]
\newtheorem{corollary}{Corollary}[section]
\theoremstyle{definition}
\newtheorem{definition}{Definition}[section]
\newtheorem{example}{Example}[section]

\DeclareMathOperator*{\argmax}{argmax}
\DeclareMathOperator*{\argmin}{argmin}

\makeatletter
\def\thanks#1{\protected@xdef\@thanks{\@thanks\protect\footnotetext{#1}}}
\makeatother

\title{\LARGE \bf \vskip -2cm Feasible strategies in three-way conflict analysis with three-valued ratings}

\author[a,b,$\ast$]{\small Jing Liu$^{\text{a,b}}$, Mengjun Hu$^{\text{c}}$, Guangming Lang}

\affil[a]{\footnotesize School of Mathematics and Statistics, Changsha University of Science and Technology, Changsha, Hunan, 410114, P.R. China}
\affil[b]{\footnotesize Hunan Provincial Key Laboratory of Mathematical Modeling and Analysis in Engineering,  Changsha University of Science and Technology, Changsha, Hunan, 410114, P.R. China}
\affil[c]{\footnotesize Department of Mathematics and Computing Science, Saint Mary’s University, Halifax, Nova Scotia, B3H 3C3, Canada}

\thanks{\hspace{-0.24cm} $^\ast$Corresponding author.}
\thanks{E-mail addresses: 
liujing19990215@163.com (Jing Liu), 
mengjun.hu@smu.ca (Mengjun Hu), 
langguangming1984@126.com (Guangming Lang).}

\date{}

\begin{document}

\baselineskip 18pt
\maketitle

\vskip -0.7cm

\noindent {\bf Abstract:} 
Most existing work on three-way conflict analysis has focused on trisecting agent pairs, agents, or issues, {\color{black}which contributes to understanding the nature of conflicts but falls short in addressing their resolution. Specifically, the formulation of feasible strategies, as an essential component of conflict resolution and mitigation, has received insufficient scholarly attention.} Therefore, this paper aims to investigate feasible strategies from two perspectives of consistency and non-consistency. Particularly, we begin with computing the overall rating of a clique of agents based on positive and negative similarity degrees. Afterwards, considering the weights of both agents and issues, we propose weighted consistency and non-consistency measures, which are respectively used to identify the feasible strategies for a clique of agents. Algorithms are developed to identify feasible strategies, $L$-order feasible strategies, and the corresponding optimal ones. Finally, to demonstrate the practicality, effectiveness, and superiority of the proposed models, we apply them to two commonly used case studies on NBA labor negotiations and development plans for Gansu Province and conduct a sensitivity analysis on parameters and a comparative analysis with existing state-of-the-art conflict analysis approaches. {\color{black}The comparison results demonstrate that our conflict resolution models outperform the conventional approaches by unifying weighted agent-issue evaluation with consistency and non-consistency measures to enable the systematic identification of not only feasible strategies but also optimal solutions.}


\section{Introduction}\label{1jie}\hspace{0.2in}

With societal progress, conflicts have become more widespread and intricate across various levels of individual, group, and organizational, where agents often exhibit opposing attitudes or behaviors due to differences in goals, resources, interests, and other factors. Conflict analysis has emerged as a critical field, offering a robust framework for understanding and resolving conflicts~\cite{Pawlak1984,Pawlak1998,Pawlak2005,Deja1996,Deja2002,R2007,Z1985,Hipel1991,Sun2015,Yu2015}. Early works by Pawlak~\cite{Pawlak1984,Pawlak1998} introduced the concept of evaluating relationships between two agents as alliance, neutrality, or conflict using auxiliary functions, laying the foundation for conflict analysis using rough set theory. Deja~\cite{Deja2002} expanded on Pawlak’s model by integrating Boolean reasoning, thereby offering a deeper understanding of conflict and addressing three core questions related to conflict resolution.

The theory of three-way decision~\cite{Yao2010,Yao2016,Yao2018,Yao2023_Dao} is closely aligned with human cognitive processes, facilitating thinking, problem-solving, and information processing by conceptualizing decisions in terms of three elements. The Triading-Acting-Optimizing model~\cite{Yao2023_Dao} presents a framework for three-way decision, which involves a Triading step for forming a triad of three elements, an Acting step for applying distinct strategies to process these elements, and an Optimizing step to evaluate and refine the overall outcome. In a more focused sense, the philosophy and methodology of three-way decision have been applied to various specific topics~\cite{Wang2022,Zhan2022,Qi2024,Yang2024,Liu2024,Cai_2024_1,Dai2024,Deng2023,WangJJ2022,WangWJ2023,Zhu2023}. Three-way conflict analysis, as one of these topics, seeks to reinforce the application of three-way decision and conflict analysis, supported by their shared foundations in decision-making and strategy formulation. {\color{black}Furthermore, various models of conflict analysis have also been investigated, including formal concept-based conflict analysis \cite{Zhi2022,Zhi2024}, reduction-combined conflict analysis \cite{Zhang2025,Chen2025}, conflict analysis with agent-agent mutual selection \cite{Dou2024}, preference-based conflict analysis \cite{Hu2025}, and multi-scale and multi-source conflict analysis \cite{Lu2025,Tang2025}.}
Current literature has explored three-way conflict analysis from two main perspectives: trisections regarding agents and issues~\cite{Lang2017,Yao2019,Xu2024,Lang2020,Li2021,Li2022,Zhi2020}, corresponding to the Triading step, and feasible strategies~\cite{Sun2016,Sun2020,Lang2020-KBS,Xu2022,Du2022,Yang2023,Li2023,Hu2023,Mandal2023}, corresponding to the Acting step.

The literature highlights several important trisections in conflict analysis. The first is the trisection of agent pairs, representing three types of agent relations. Lang, Miao, and Cai~\cite{Lang2017} integrated three-way decision theory and decision-theoretic rough sets into conflict analysis, defining probabilistic conflict, neutrality, and alliance relations, which form a trisection of agent pairs. They used decision-theoretic rough sets to compute thresholds for the trisection. Yao~\cite{Yao2019} introduced a distance function to analyze the trisection of agent pairs, categorizing conflicts into strong-, weak-, and non-conflict levels. Xu and Jia~\cite{Xu2024} applied similarity functions to analyze the trisection of agent pairs in the context of single and multiple issues and proposed a method for selecting optimal trisection thresholds based on a measure. The second is the trisection of individual agents, representing their alliances or groups. Lang, Miao, and Fujita~\cite{Lang2020} explored three-way group conflict using Pythagorean fuzzy information, defining positive, neutral, and negative alliances based on fuzzy loss functions and Bayesian minimal risk theory. Li et al.~\cite{Li2021} applied triangular fuzzy situation tables to evaluate agent attitudes toward multiple issues, using decision-theoretic rough sets to calculate two thresholds for the trisection. In a subsequent study~\cite{Li2022}, they extended this analysis to multiple issues with trapezoidal fuzzy situation tables. The third is the trisection of individual issues. Zhi et al.~\cite{Zhi2020} proposed a novel conflict analysis framework based on three-way concept analysis, particularly for conflict situations involving one-vote veto. They identified the maximum alliance and minimum conflict sets within cliques based on the allied, conflict, and neutral issues. {\color{black}While these studies successfully employ trisections to uncover the fundamental nature of conflicts, the formulation of practical solutions remains the ultimate objective. Therefore, the immediate priority is to explore feasible strategies for effectively resolving conflict problems.}

To construct feasible strategies, Sun, Ma, and Zhao~\cite{Sun2016} introduced a matrix-based approach grounded in rough sets over two universes to identify the root causes of conflict and derive the maximum feasible consensus strategies in conflict situations. They later extended their model into a probabilistic context~\cite{Sun2020}. Xu et al.~\cite{Xu2022} developed consistency measures to assess the similarity between agents' attitudes and consensus attitudes, defining $L$-order dominant feasible strategies based on the consistency degree of a clique concerning multiple issues.
{\color{black} Although these studies investigated the identification of feasible strategies, they failed to consider the relative importance of agents and issues. Moreover, Xu's work did not address the determination of optimal feasible strategies.}
Du et al.~\cite{Du2022} examined three types of relations between agent pairs from the perspective of absolute and relative conflicts, considering all subsets of the union set of weak-conflict and non-conflict sets as feasible strategies. They proposed an evaluation function to assess these strategies. Similarly, Yang, Wang, and Guo~\cite{Yang2023} investigated three-way conflict analysis and resolution within hybrid situation tables, while Li, Qiao, and Ding~\cite{Li2023} conducted conflict analysis using $q$-rung orthopair fuzzy information to select feasible strategies. 
{\color{black} Although above scholars investigated both feasible strategies and optimal ones, their approaches only incorporated issue weights while neglecting the weights of agents.}

While trisections have received significant attention, there remains a notable gap in the exploration of feasible strategies. In this work, we aim to advance Xu's model with consistency measures~\cite{Xu2022}, addressing several weaknesses in their strategy selection method. {\color{black} Specifically, this study incorporates agent and issue weights to provide greater flexibility in evaluating the overall ratings of agent groups and in refining the selection of diverse strategy types for conflict resolution. In addition, it introduces a novel perspective by employing non-consistency measures for strategy selection, thereby offering a complementary approach to traditional consistency-based methods.} Our main contributions are as follows:
\begin{enumerate}[label = (\arabic*)]
\setlength{\itemsep}{0pt}
\item \textbf{A revised approach to selecting feasible strategies based on consistency measures that integrate both agent and issue weights}: We propose the weighted consistency measures and demonstrate that they extend and advance those proposed by Xu~\cite{Xu2022} in enabling more flexible and practical modeling of conflict and strategy selection.

\item \textbf{A novel approach to selecting feasible strategies based on non-consistency measures}: We propose two non-consistency measures that incorporate agent and issue weights to guide strategy selection, offering an alternative to consistency measures in understanding, formulating, and resolving conflict. 

\item \textbf{Algorithms for identifying feasible strategies with both consistency and non-consistency measures}: We present algorithms for both the consistency-based and non-consistency-based approaches, demonstrated {\color{black} through case studies of NBA labor negotiations and development plans of Gansu Province.} A comparative analysis of our proposed methods against existing relevant models is also implemented. 
\end{enumerate}

\begin{figure}[!ht]
\centering
\includegraphics[width=6.3in]{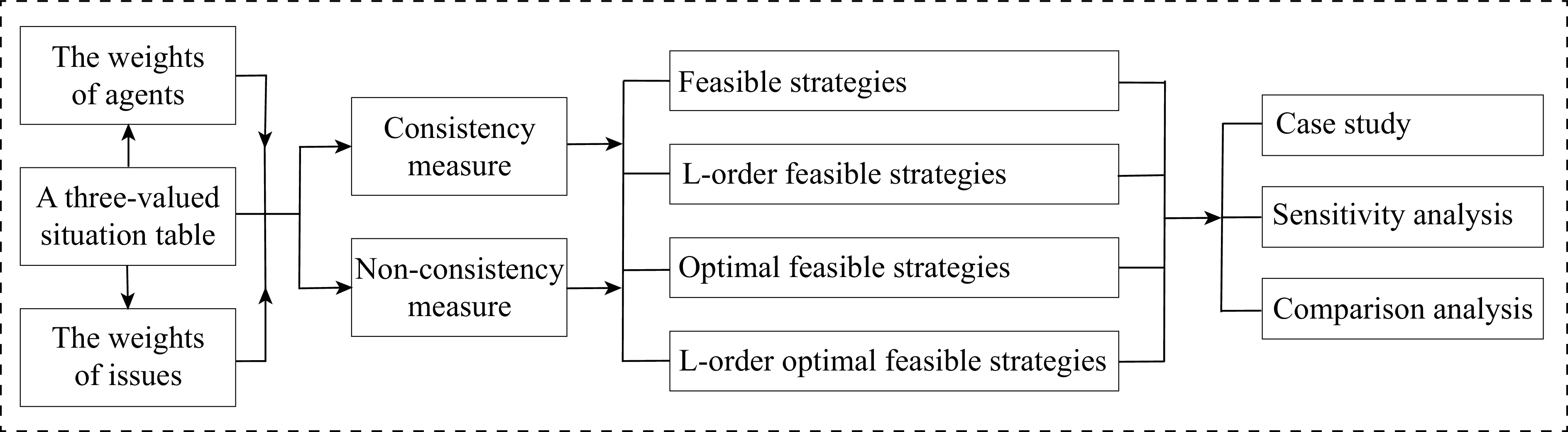}
\caption{The flowchart of  constructing feasible strategies.}
\label{fig_1}
\end{figure}

The remainder of this paper is structured as follows. Section~\ref{2jie} provides a brief overview of Xu's model with consistency measures. Section~\ref{3jie} presents our approach using consistency measures, while Section~\ref{4jie} introduces our approach using non-consistency measures. Algorithms for both approaches are detailed in these sections. Two case studies, a sensitivity analysis, and a comparative analysis are performed in Section~\ref{5jie}. Finally, Section~\ref{6jie} concludes the paper and discusses future research directions. {\color{black}The flowchart of this paper is presented in Figure \ref{fig_1}.}

\section{A review of three-way conflict analysis with consistency measures}\label{2jie}\hspace{0.2in}
In this section, we briefly review a model of three-way conflict analysis based on consistency measures proposed by Xu \cite{Xu2022}, with a few reformulations. Specifically, the model focuses on a three-valued situation table defined as follows.

\begin{definition}\textit{(Yao \cite{Yao2019}, 2019)}
A three-valued situation table is formally represented by a quadruple $TS = (P, T, V, r)$, where $P = \{p_1, p_2, \dots, p_m\}$ is a non-empty finite set of agents, $T = \{t_1, t_2, \dots, t_n\}$ is a non-empty finite set of issues, $V = \{-1, 0, +1\}$ is the set of three-valued ratings, and $r: P \times T \to V$ is a rating function that assigns a value from $V$ to each agent-issue pair. For an agent $p \in P$ and an issue $t \in T$, the rating $r(p,t) = +1$ indicates that the agent $p$ supports the issue $t$; $r(p,t) = 0$ indicates that $p$ is neutral to $t$; and $r(p,t) = -1$ indicates that $p$ opposes $t$.
\end{definition}

All following definitions and arguments are discussed based on a three-way situation table $TS = (P, T, V, r)$, which will not be reiterated and highlighted in subsequent discussions.

\begin{example}\label{ex2.1}
Table \ref{tab_middle-east} presents the Middle East conflict situation as the initial example of a three-valued situation table introduced by Pawlak \cite{Pawlak1998}. The agents $p_1, p_2, p_3, p_4, p_5,$ and $p_6$ represent Israel, Egypt, Palestine, Jordan, Syria, and Saudi Arabia, respectively. The issues $t_1, t_2, t_3, t_4,$ and $t_5$ are, respectively: ``Autonomous Palestinian state on the West Bank and Gaza'', ``Israeli military outposts along the Jordan River'', ``Israel retains East Jerusalem'', ``Israeli military outposts on the Golan Heights'', and ``Arab countries grant citizenship to Palestinians who choose to remain within their borders.'' In Table \ref{tab_middle-east}, for simplicity, the ratings ``$+1$'' and ``$-1$'' are omitted as ``$+$'' and ``$-$'', respectively.

\begin{table}[!ht]
\renewcommand{\arraystretch}{1.3}
\begin{center}
\caption{The Middle East conflict situation table~\cite{Pawlak1998}.}\label{tab_middle-east}
\setlength{\tabcolsep}{11.8mm}
\begin{tabular}{cccccc}
\hline
 & $t_{1}$ & $t_{2}$ & $t_{3}$ & $t_{4}$ & $t_{5}$ \\
\hline
$p_{1}$ & $-$ & $+$ & $+$ & $+$ & $+$ \\
$p_{2}$ & $+$ & $0$ & $-$ & $-$ & $-$ \\
$p_{3}$ & $+$ & $-$ & $-$ & $-$ & $0$ \\
$p_{4}$ & $0$ & $-$ & $-$ & $0$ & $-$ \\
$p_{5}$ & $+$ & $-$ & $-$ & $-$ & $-$ \\
$p_{6}$ & $0$ & $+$ & $-$ & $0$ & $+$ \\
\hline
\end{tabular}
\end{center}
\end{table}
\end{example}

While the rating function $r$ specifies the raw attitudes of individual agents on individual issues, there are diverse opinions and approaches when synthesizing them regarding a group of agents (i.e. a clique), a group of issues, or both. Xu et al. \cite{Xu2022} presented an approach that implicitly assumes two ``imaginary'' agents: a positive ideal agent supporting all issues and a negative ideal agent opposing all issues. We formally define these two agents to make their usage explicit.

\begin{definition}
The positive ideal agent $p^{+}$ and the negative ideal agent $p^{-}$ are two imaginary agents satisfying the following conditions respectively: $\forall t\in T, r(p^{+}, t) = +1$ and $\forall t\in T, r(p^{-}, t) = -1$. 
\end{definition}

These two ideal agents represent two opposite extremes, analogous to the positive and negative ideal solutions in the TOPSIS method \cite{Lai1994}. With $p^{+}$ and $p^{-}$ as references at two extremes, we can measure their similarity with a group of agents, which indicates how positive or negative the group as a whole is. Specifically, the similarity measure presented by Xu \cite{Xu2022} can be reformulated with an explicit use of $p^{+}$ and $p^{-}$ as follows.

\begin{definition}
For an issue $t \in T$ and a clique of agents $G \subseteq P$, the positive similarity degree $\mathbb{S}_{t}^{+}(G)$ between $G$ and the positive ideal agent $p^{+}$ on $t$ and the negative similarity degree $\mathbb{S}_{t}^{-}(G)$ between $G$ and the negative ideal agent $p^{-}$ on $t$ are defined as: 
\begin{align}\label{eq_S-Xu}
\mathbb{S}_{t}^{+}(G) &= 1 - \frac{\sum\limits_{p \in G} \left| r(p,t) - r(p^{+}, t) \right|}{2\#(G)}, \notag \\
\mathbb{S}_{t}^{-}(G) &= 1 - \frac{\sum\limits_{p \in G} \left| r(p,t) - r(p^{-}, t) \right|}{2\#(G)}, 
\end{align}
where $\left| \cdot \right|$ represents the absolute value of a number, and $\#(\cdot)$ the cardinality of a set.
\end{definition}

The positive similarity degree $\mathbb{S}_{t}^{+}(G)$ signifies the degree of consistency between the clique $G$ and the positive ideal agent $p^{+}$ concerning the issue $t$; the negative similarity degree $\mathbb{S}_{t}^{-}(G)$ indicates the degree of consistency between the clique $G$ and the negative ideal agent $p^{-}$ regarding the issue $t$.

A comparison of $\mathbb{S}_{t}^{+}(G)$ and $\mathbb{S}_{t}^{-}(G)$ helps understand the overall attitude of the clique $G$ on $t$. While Xu et al. \cite{Xu2022} called it the consensus attitude, we adopt a more unified formulation by extending the rating function $r$. Specifically, we generalize $r$ regarding an individual agent and an individual issue to a rating function $R: 2^{P} \times T$ regarding a clique of agents and an individual issue. Intuitively, $R(G,t)$ represents the overall/consensus rating of $G$ on $t$. Although not considered in this work, similar rating formulations can be defined on $P \times 2^{T}$ signifying the overall rating of an individual agent on an issue set and $2^{P} \times 2^{T}$ indicating the overall rating of a clique of agents on an issue set. These generalizations provide a unified formulation of various types of consensus ratings. They may also inspire the connections between conflict analysis and granular computing.

With a qualitative comparison of $\mathbb{S}_{t}^{+}(G)$ and $\mathbb{S}_{t}^{-}(G)$, Xu et al. \cite{Xu2022} considered a three-valued overall rating of $G$ on $t$, which is reformulated through a rating function on $2^{P} \times T$ as follows.

\begin{definition}\label{def_R-Xu}
The rating for a clique of agents $G \subseteq P$ on an individual issue $t \in T$ is defined by a function $R^{Xu}: 2^{P} \times T \rightarrow \{+1,0,-1\}$ as: 
\begin{align}\label{eq_R-Xu}
R^{Xu}(G,t) = \left\{
\begin{array}{lcl}
+1,  && \text{if~~} \mathbb{S}_{t}^{+}(G) > \mathbb{S}_{t}^{-}(G), \vspace{2mm} \\
0,   && \text{if~~} \mathbb{S}_{t}^{+}(G) = \mathbb{S}_{t}^{-}(G), \vspace{2mm} \\
-1,  && \text{if~~} \mathbb{S}_{t}^{+}(G) < \mathbb{S}_{t}^{-}(G).
\end{array}
\right.
\end{align}
\end{definition}

We use the superscript to distinguish it from the function proposed in our work. With the overall rating $R^{Xu}(G,t)$ as a reference, Xu et al. \cite{Xu2022} measured the consistency within $G$ on $t$ through the distances between an individual agent's rating $r(p,t)$ and the clique's overall rating $R^{Xu}(G,t)$, equivalently reformulated as follows using $R^{Xu}$.

\begin{definition}\label{def_CMi-Xu}
The consistency measure $\mathbb{CM}_{t}^{Xu}: 2^{P} \to [0.5, 1]$ regarding an issue $t \in T$ is defined as: for a clique of agents $G \subseteq P$, 
\begin{align}\label{eq_CMi-Xu}
\mathbb{CM}_{t}^{Xu}(G) = 1 - \frac{\sum\limits_{p \in G} \left| r(p,t) - R^{Xu}(G,t) \right|}{2\#(G)}.
\end{align}
\end{definition}

When $\mathbb{CM}_{t}^{Xu}(G) = 1$, the attitudes from individual agents in the clique $G$ regarding the issue $t$ are fully consistent. In other words, all agents in $G$ hold the same attitude, either positive, neutral, or negative, on $t$. Xu et al. \cite{Xu2022} further defined a consistency measure regarding an issue set $J\subseteq T$ by taking an average over individual issues in $J$.

\begin{definition}\textit{(Xu \cite{Xu2022}, 2022)}\label{def_CMJ-Xu}
The consistency measure $\mathbb{CM}_{J}^{Xu}: 2^{P} \to [0.5, 1]$ concerning a non-empty issue set $J \subseteq T$ is defined as: for a clique of agents $G \subseteq P$, 
\begin{align}\label{eq_CMJ-Xu}
\mathbb{CM}_{J}^{Xu}(G) = \frac{\sum\limits_{t \in J} \mathbb{CM}_{t}^{Xu}(G)}{\#(J)}.
\end{align}
\end{definition}

By the definition, the attitudes of individual agents within $G$ are fully consistent regarding $J$ if and only if their attitudes are fully consistent on all individual issues in $J$.

The consistency measure $\mathbb{CM}_{t}^{Xu}$ can also be used as an evaluation for an individual issue $t$. Accordingly, one can rank the issue set $T = \{t_{1}, t_{2}, \dots, t_{n}\}$ as $T = \{t'_{1}, t'_{2}, \dots, t'_{n}\}$ regarding the clique $G \subseteq P$, in the descending order of consistency. Based on this ranking, Xu et al. \cite{Xu2022} defined the $L$-order dominant feasible strategy for the clique $G$.

\begin{definition}\textit{(Xu \cite{Xu2022}, 2022)}\label{def_FS-Xu}
For a clique of agents $G \subseteq P$ and $T=\{t'_{1}, t'_{2}, \dots , t'_{n}\}$ with $\mathbb{CM}_{t'_{1}}^{Xu}(G) \geq \mathbb{CM}_{t'_{2}}^{Xu}(G) \geq \cdots \geq \mathbb{CM}^{Xu}_{t'_{n}}(G)$. The set $FS_{L}^{Xu}(G) = \{t'_1, t'_2, \dots, t'_L\}~(1 \leq L \leq n)$ is called an $L$-order dominant feasible strategy for $G$ if: 
\begin{align}\label{eq_FS-Xu}
\frac{\#(G)}{\#(P)} \geq \gamma_P \wedge \mathbb{CM}^{Xu}_{FS_{L}^{Xu}(G)}(G) \geq \lambda, 
\end{align} 
where $\gamma_P\in [0,1]$ and $\lambda\in [0.5,1]$ are two given thresholds.
\end{definition}

Equation \eqref{eq_FS-Xu} states two conditions: (1) the clique $G$ must be large enough; (2) the overall consistency degree within $G$ on the top-ranked $L$ issues must be large enough.

While the above consistency measures find their usefulness in certain scenarios, there are a few limitations that can be addressed to make them more practical in real-world applications: 
\begin{enumerate}[label = (\arabic*)]
\setlength{\itemsep}{0pt}
\item The consistency measure $\mathbb{CM}_{J}^{Xu}$ regarding a set of issues (i.e. Definition \ref{def_CMJ-Xu}) is central to constructing the $L$-order dominant feasible strategy. One possible improvement is to contemplate the weights of agents and issues. Agent weights represent the relative powers of agents, and issue weights reflect the relative importance of the issues. For example, in a company, shareholders hold varying numbers of shares, which leads to their different powers during a general meeting. Moreover, the issues discussed in the meeting may have varying degrees of impact on the company, indicating differences in their importance. By incorporating the weights of both agents and issues into the consistency measure, conflict can be quantified more accurately and resolved more effectively. 

\item The overall rating of a clique is a critical factor in defining $\mathbb{CM}_{t}^{Xu}$ and accordingly in constructing the $L$-order dominant feasible strategy. Definition \ref{def_R-Xu} considers which of $\mathbb{S}_{t}^{+}(G)$ and $\mathbb{S}_{t}^{-}(G)$ is larger. As a result, $G$ is determined to be positive or negative on $t$ even if $\mathbb{S}_{t}^{+}(G)$ and $\mathbb{S}_{t}^{-}(G)$ are close (e.g. $\mathbb{S}_{t}^{+}(G)=0.5001$ and $\mathbb{S}_{t}^{-}(G)=0.4999$). Accordingly, a possible improvement is to count in how significant the difference between $\mathbb{S}_{t}^{+}(G)$ and $\mathbb{S}_{t}^{-}(G)$ is. Intuitively, $G$ is overall positive on $t$ if and only if $\mathbb{S}_{t}^{+}(G)$ is substantially larger than $\mathbb{S}_{t}^{-}(G)$. Similarly, $G$ is overall negative on $t$ if and only if $\mathbb{S}_{t}^{-}(G)$ is substantially larger than $\mathbb{S}_{t}^{+}(G)$. Otherwise, $\mathbb{S}_{t}^{+}(G)$ and $\mathbb{S}_{t}^{-}(G)$ are close or equal, and $G$ is overall neutral on $t$.

\item A feasible strategy refers to a course of actions that allows as many agents as possible to reach an agreement on as many issues as possible, within acceptable limits. One important objective in conflict analysis is to identify feasible strategies and optimal ones for resolving conflicts. In practice, multiple feasible strategies often exist in a conflict situation. With the consistency measures proposed by Xu \cite{Xu2022}, one can select $L$-order dominant feasible strategies based on a fixed ranking of issues. However, how to determine the optimal feasible strategy is not explored. 
\end{enumerate}

To address the above limitations, this work introduces weighted consistency and non-consistency measures that consider the significance of the weights of both agents and issues and discusses the selection of optimal feasible strategies. The next two sections present the approaches based on the consistency and non-consistency measures, respectively.

\section{Constructing feasible strategies with consistency measures}\label{3jie}\hspace{0.2in}
In this section, we propose two consistency measures considering the weights of both agents and issues, based on which, we demonstrate how to construct feasible strategies and accordingly, how to select the optimal strategies.

\subsection{Weights of agents and issues}\label{3.1jie}\hspace{0.2in}
We formally discuss a few different types of weights regarding agents and issues. Specifically, we contemplate the following four cases of weights regarding agents: 
\begin{enumerate}[label = (\arabic*)]
\setlength{\itemsep}{0pt}
\item $\theta(p)$: the weight of an individual agent $p$ relative to $P$. \\
This weight is assumed to be given as an input to the model. It quantifies the power of $p$ in the universal group of agents. It can be alternatively denoted as $\theta(p|P)$. Although not necessary, we assume $\sum\limits_{p \in P}\theta(p)=1$.

\item $\theta(G)$: the weight of a clique of agents $G \subseteq P$ relative to $P$. \\
This weight represents the overall power of $G$ as a group in the context of $P$. Intuitively, it is determined by the power of individual agents in $G$ and can be computed in a variety of ways. In this work, we take a simple sum, that is: 
\begin{align}\label{eq_theta-X}
\theta(G)=\sum\limits_{p \in G}\theta(p).
\end{align}
It can be alternatively denoted as $\theta(G|P)$. 

\item $\theta(p|G)$: the (conditional) weight of an individual agent $p$ relative to a clique $G$. \\
This weight focuses on the power of $p$ within a specific clique $G$. One way to compute it is: 
\begin{align}\label{eq_theta-aX}
\theta(p|G) = \frac{\theta(p|P)}{\theta(G|P)}.
\end{align}
Apparently, we have $\sum\limits_{p \in G}\theta(p|G) =1$ if we take Equation \eqref{eq_theta-X}.

\item $\theta(H|G)$: the (conditional) weight of a sub-clique $H \subseteq G$ relative to $G$. \\
This weight looks at the power of a sub-group $H$ in the context of $G$. It is useful to compare the powers among different coalitions or groups within a certain bigger clique. In this work, we again simply accumulate the power of individual agents:
\begin{align}\label{eq_YX}
\theta(H|G) = \sum\limits_{p \in H}\theta(p|G).
\end{align}
\end{enumerate}

The weights regarding issues can be similarly formulated. We omit the detailed discussion for simplicity. Intuitively, one may consider the weights of individual issues or issue sets relative to the universe $T$ or a specific issue set $J\subseteq T$. For distinguishing purposes, we will use $\omega$, in place of $\theta$, to denote the weights regarding issues.

For a clique $G \subseteq P$, one may get three straightforward coalitions according to the agents' raw attitudes towards a specific issue.

\begin{definition}\label{def_tri-Xi}
The positive coalition $G_{t}^{+}$, neutral coalition $G_{t}^{0}$, and negative coalition $G_{t}^{-}$ of a clique $G \subseteq P$ regarding an issue $t \in T$ are defined as: 
\begin{align}
G_{t}^{+} &= \{p \in G \mid r(p,t) = +1\}, \notag\\
G_{t}^{0} &= \{p \in G \mid r(p,t) = 0\},  \label{eq_Xi}\\
G_{t}^{-} &= \{p \in G \mid r(p,t) = -1\}. \notag
\end{align}
\end{definition}

The three coalitions form a weak tri-partition of $G$, that is, 
$G_{t}^{+} \cap G_{t}^{0} = \emptyset, G_{t}^{+} \cap G_{t}^{-} = \emptyset, G_{t}^{-} \cap G_{t}^{0} = \emptyset$, 
and $G_{t}^{+} \cup G_{t}^{0} \cup G_{t}^{-} = G$.
Each coalition represents a unique attitude, and the weight of each group reflects the relative power of $G$ on the corresponding attitude. For instance, $\theta(G_{t}^{+}|G)$ quantifies the positive power on the issue $t$ within $G$.

\begin{definition}\label{def_rho}
The positive power $\rho_{t}^{+}(G)$, neutral power $\rho_{t}^{0}(G)$, and negative power $\rho_{t}^{-}(G)$ of a clique $G \subseteq P$ regarding an issue $t \in T$ are defined as:
\begin{align}
\rho_{t}^{+}(G) &= \theta(G_{t}^{+}|G) = \sum\limits_{p \in G_{t}^{+}} \theta(p|G) 
= \sum\limits_{p \in G_{t}^{+}} \frac{\theta(p)}{\theta(G)}, \notag\\
\rho_{t}^{0}(G) &= \theta(G_{t}^{0}|G) = \sum\limits_{p \in G_{t}^{0}} \theta(p|G) 
= \sum\limits_{p \in G_{t}^{0}} \frac{\theta(p)}{\theta(G)}, \label{eq_rho}\displaybreak[1]\\
\rho_{t}^{-}(G) &= \theta(G_{t}^{-}|G) = \sum\limits_{p \in G_{t}^{-}} \theta(p|G) 
= \sum\limits_{p \in G_{t}^{-}} \frac{\theta(p)}{\theta(G)}. \notag
\end{align}
\end{definition}

In the special case where agents are given equal power in $P$, we have: 
\begin{align}
\rho_{t}^{+}(G) = \frac{\#(G_{t}^{+})}{\#(G)}, \quad
\rho_{t}^{0}(G) = \frac{\#(G_{t}^{0})}{\#(G)}, \quad
\rho_{t}^{-}(G) = \frac{\#(G_{t}^{-})}{\#(G)}, 
\end{align}
which are totally determined by their cardinalities. This is also the case where weights are not considered.

By Definition \ref{def_rho}, we can easily get the following proposition.

\begin{proposition}\label{proposition_rho}
For a clique $G \subseteq P$ and an issue $t \in T$, we have: 
\begin{enumerate}[label = (\arabic*),font = \normalfont]
\setlength{\itemsep}{0pt}
\item $\rho_{t}^{+}(G),\rho_{t}^{0}(G),\rho_{t}^{-}(G)\in[0,1];$
\item $\rho_{t}^{+}(G)+\rho_{t}^{0}(G)+\rho_{t}^{-}(G)=1$.
\end{enumerate}
\end{proposition}

\begin{example}\textbf{(Continuing with Example \ref{ex2.1})}\label{ex3.1}
This example illustrates the calculation of the positive, neutral, and negative powers of a clique concerning a single issue.
{\color{black}The weight vector of all agents is provided as $\Theta = (\theta(p_1), \theta(p_2), \dots, \theta(p_6))= (0.25, 0.18, 0.1, 0.15, 0.12, 0.2)$ based on expert experience.} Firstly, for the clique $G = \{p_1, p_3, p_4, p_6\}$, the weights of agents with respect to $G$ are calculated as:
\begin{align*}
\theta(p_1 | G) &= \frac{0.25}{0.25 + 0.1 + 0.15 + 0.2} = \frac{5}{14}, \\
\theta(p_3 | G) &= \frac{0.1}{0.25 + 0.1 + 0.15 + 0.2} = \frac{1}{7}, \\
\theta(p_4 | G) &= \frac{0.15}{0.25 + 0.1 + 0.15 + 0.2} = \frac{3}{14}, \\
\theta(p_6 | G) &= \frac{0.2}{0.25 + 0.1 + 0.15 + 0.2} = \frac{2}{7}.
\end{align*}
Secondly, by Definition~\ref{def_tri-Xi}, we have the following coalitions of $G$ regarding the issue $t_1$:
\begin{align*}
G_{t_1}^{+} = \{p_3\}, \quad
G_{t_1}^{0} = \{p_4, p_6\}, \quad
G_{t_1}^{-} = \{p_1\}.
\end{align*}
Finally, by Definition~\ref{def_rho}, the positive, neutral, and negative powers of $G$ on $t_1$ are:
\begin{align*}
\rho_{t_1}^{+}(G) &= \theta(p_3 | G) = \frac{1}{7}, \\
\rho_{t_1}^{0}(G) &= \theta(p_4 | G) + \theta(p_6 | G) = \frac{1}{2}, \\
\rho_{t_1}^{-}(G) &= \theta(p_1 | G) = \frac{5}{14}.
\end{align*}
Similarly, for the remaining issues, we can obtain the following powers:
\begin{align*}
&\rho_{t_2}^{+}(G) = \frac{9}{14}, \qquad \rho_{t_2}^{0}(G) = 0, \qquad~ \rho_{t_2}^{-}(G) = \frac{5}{14}; \\
&\rho_{t_3}^{+}(G) = \frac{5}{14}, \qquad \rho_{t_3}^{0}(G) = 0, \qquad~ \rho_{t_3}^{-}(G) = \frac{9}{14}; \displaybreak[1] \\
&\rho_{t_4}^{+}(G) = \frac{5}{14}, \qquad \rho_{t_4}^{0}(G) = \frac{1}{2}, \qquad \rho_{t_4}^{-}(G) = \frac{1}{7}; \\
&\rho_{t_5}^{+}(G) = \frac{9}{14}, \qquad \rho_{t_5}^{0}(G) = \frac{1}{7}, \qquad \rho_{t_5}^{-}(G) = \frac{3}{14}.
\end{align*}
\end{example}

\subsection{Consistency measures incorporating the weights of agents and issues}\label{3.2jie}\hspace{0.2in}
In this section, we present two consistency measures that integrate both agent and issue weights, respectively regarding a single issue and a set of issues.

We first focus on the consistency measure regarding a single issue. Following the intuitive idea by Xu \cite{Xu2022}, reviewed in Section \ref{2jie}, the consistency within a clique $G \subseteq P$ on an issue $t \in T$ can be measured through the distances between an individual agent's rating on $t$, given by $r(p,t)$, and a clique's overall rating on $t$, that is, $R(G,t)$. This overall rating can be determined by a comparison of positive and negative degrees of $G$. These two degrees can be further decomposed with respect to individual agents, measured by their similarity to the two ideal agents $p^{+}$ and $p^{-}$.

\begin{definition}
The positive degree of an agent on an issue is defined by a function $\mathbb{SA}^{+}: P \times T \rightarrow [0,1]$, measured by the similarity between the agent and the positive ideal agent $p^{+}$ as:
\begin{align}\label{eq_SA+ai}
\mathbb{SA}^{+}(p,t) = 1 - \frac{|r(p, t) - r(p^{+}, t)|}{2}.
\end{align}
Similarly, the negative degree is defined by a function $\mathbb{SA}^{-}: P \times T \rightarrow [0,1]$, measured by the similarity between the agent and the negative ideal agent $p^{-}$ as:
\begin{align}\label{eq_SA-ai}
\mathbb{SA}^{-}(p,t) = 1 - \frac{|r(p, t) - r(p^{-}, t)|}{2}. 
\end{align}
\end{definition}

Differently from Xu's work, we put the issue as a parameter of the functions rather than defining separate functions for issues. This facilitates a uniform formulation for the positive and negative degrees. The same idea will be applied to other concepts in our work.

As we have $-1 = r(p^{-},t) \leq r(p,t) \leq r(p^{+},t) = 1$ for any agent $p$ and issue $t$, we can obtain the following properties. The proof is straightforward and thus, omitted.

\begin{proposition}\label{proposition_SA+-ai}
For an agent $p \in P$ and an issue $t \in T$, we have:
\begin{enumerate}[label = (\arabic*),font = \normalfont]
\setlength{\itemsep}{0pt}
\item $\mathbb{SA}^{+}(p,t) = \frac{1+r(p,t)}{2}$, 
$\mathbb{SA}^{-}(p,t) = \frac{1-r(p,t)}{2};$
\item $\mathbb{SA}^{+}(p,t) + \mathbb{SA}^{-}(p,t) = 1;$
\item $\mathbb{SA}^{+}(p,t) - \mathbb{SA}^{-}(p,t) = r(p,t)$.
\end{enumerate}
\end{proposition}

The positive and negative degrees of a clique can be measured by aggregating the degrees of individual agents, taking their weights/power into account.

\begin{definition}\label{def_SA+-Xi}
The positive degree of a clique on an issue is defined by a function $\mathbb{SA}^{+\prime}: 2^{P} \times T \rightarrow [0,1]$, measured by the weighted similarity between the clique and the positive ideal agent $p^{+}$ as:
\begin{align}\label{eq_SA+Xi}
\mathbb{SA}^{+}(G,t) = \sum_{p \in G} \big( \theta(p|G) \times \mathbb{SA}^{+}(p,t) \big).
\end{align} 
Similarly, the negative degree is defined by a function $\mathbb{SA}^{-\prime}: 2^{P} \times T \rightarrow [0,1]$, measured by the weighted similarity between the clique and the negative ideal agent $p^{-}$ as:
\begin{align}\label{eq_SA-Xi}
\mathbb{SA}^{-}(G,t) = \sum_{p \in G} \big( \theta(p|G) \times \mathbb{SA}^{-}(p,t) \big).
\end{align}
\end{definition}

For simplicity, we will omit the prime sign in the superscript of $\mathbb{SA}^{+\prime}$ and $\mathbb{SA}^{-\prime}$ if the types of parameters are not confusing.

In the case where the weights of agents are equal or not considered, we have $\theta(p|G) = \frac{1}{\#(G)}$ for every agent $p$ in $G$. Consequently, $\mathbb{SA}^{+}(G,t)$ and $\mathbb{SA}^{-}(G,t)$ are computed as:
\begin{align}
\mathbb{SA}^{+}(G,t) &=  1 - \frac{\sum\limits_{p \in G} \left| r(p,t) - r(p^{+},t) \right|}{2\#(G)}, \notag \\
\mathbb{SA}^{-}(G,t) &=  1 - \frac{\sum\limits_{p \in G} \left| r(p,t) - r(p^{-},t) \right|}{2\#(G)}.
\end{align}
That is, they degenerate to the positive and negative similarity degrees in Equation \eqref{eq_S-Xu} defined by Xu \cite{Xu2022}.

By Definition \ref{def_SA+-Xi}, we can get the following proposition on the properties of $\mathbb{SA}^{+}(G,t)$ and $\mathbb{SA}^{-}(G,t)$.

\begin{proposition}\label{proposition_SA+-X}
For a clique $G \subseteq P$ and an issue $t \in T$, the following properties hold:
\begin{enumerate}[label = (\arabic*), font = \normalfont]
\setlength{\itemsep}{0pt}
\item $\mathbb{SA}^{+}(G,t)\in [0,1], \mathbb{SA}^{-}(G,t) \in [0,1];$
\item $\mathbb{SA}^{+}(G,t) = 1 \Leftrightarrow \forall p \in G, r(p,t) = +1$, \\
$\mathbb{SA}^{+}(G,t) = 0 \Leftrightarrow \forall p \in G, r(p,t) = -1;$
\item $\mathbb{SA}^{-}(G,t) = 1 \Leftrightarrow \forall p \in G, r(p,t) = -1$, \\
$\mathbb{SA}^{-}(G,t) = 0 \Leftrightarrow \forall p \in G, r(p,t) = +1;$
\item $\forall p \in G, r(p,t) = 0 \Rightarrow \mathbb{SA}^{+}(G,t) = \mathbb{SA}^{-}(G,t) = 0.5;$
\item $\mathbb{SA}^{+}(G,t) + \mathbb{SA}^{-}(G,t) = 1;$
\item $\mathbb{SA}^{+}(G,t) - \mathbb{SA}^{-}(G,t) = \rho_{t}^{+}(G)-\rho_{t}^{-}(G)$.
\end{enumerate}
\end{proposition}

The properties in (1)-(4) are straightforward from Definition \ref{def_SA+-Xi}. The properties in (5) and (6) can be easily verified by the properties (2) and (3) in Proposition \ref{proposition_SA+-ai}. They are actually the generalizations of those in Proposition \ref{proposition_rho} from a single agent to a clique. The proofs are omitted for simplicity.

The overall rating of a clique $G$ on issue $t$ is determined by the difference between $\mathbb{SA}^{+}(G,t)$ and $\mathbb{SA}^{-}(G,t)$. Intuitively, if $\mathbb{SA}^{+}(G,t)$ is sufficiently larger than $\mathbb{SA}^{-}(G,t)$, $G$ has an overall positive attitude on $t$; if $\mathbb{SA}^{-}(G,t)$ is sufficiently larger than $\mathbb{SA}^{+}(G,t)$, $G$ has an overall negative attitude on $t$; otherwise, $G$ is overall neutral on $t$. Accordingly, we formally define the rating function $R$ with the aid of two thresholds.

\begin{definition}\label{def_R}
For two thresholds $\mu$ and $\nu$ satisfying $-1 \leq \nu \leq 0 \leq \mu \leq 1$, the rating for a clique of agents on an individual issue is defined by a function $R: 2^{P} \times T \rightarrow \{+1,0,-1\}$ as:
\begin{align}\label{eq_At}
R(G,t) = \left\{
\begin{array}{lcl}
+1,  && \text{if~~} \mathbb{SA}^{+}(G,t) - \mathbb{SA}^{-}(G,t) > \mu, \vspace{2mm} \\
0,   && \text{if~~} \nu \leq \mathbb{SA}^{+}(G,t) - \mathbb{SA}^{-}(G,t) \leq \mu, \vspace{2mm} \\
-1,  && \text{if~~} \mathbb{SA}^{+}(G,t) - \mathbb{SA}^{-}(G,t) < \nu.
\end{array}
\right.
\end{align}
\end{definition}

Recall that in the special case where agents have equal power or are not weighted, the degrees $\mathbb{SA}^{+}(G,t)$ and $\mathbb{SA}^{-}(G,t)$ degenerate to the degrees $\mathbb{S}_{t}^{+}(G)$ and $\mathbb{S}_{t}^{-}(G)$ defined by Xu \cite{Xu2022}. Furthermore, if we take the two thresholds as $\mu = \nu = 0$, our overall rating function degenerates to Xu's version: 
\begin{align}
R(G,t) = R^{Xu}(G,t) = \left\{
\begin{array}{lcl}
+1,  && \text{if~~}  \mathbb{S}_{t}^{+}(G) > \mathbb{S}_{t}^{-}(G), \vspace{2mm} \\
0,   && \text{if~~}  \mathbb{S}_{t}^{+}(G) = \mathbb{S}_{t}^{-}(G), \vspace{2mm} \\
-1,  && \text{if~~}  \mathbb{S}_{t}^{+}(G) < \mathbb{S}_{t}^{-}(G).
\end{array}
\right.
\end{align}

In other words, Xu's approach becomes a special case of ours. Moreover, by introducing two thresholds $\mu$ and $\nu$, we address the challenge of justifying overall ratings in the existing literature, as raised in \cite{Xu2022}.

By Definition \ref{def_R} and Proposition \ref{proposition_SA+-X}, the conditions in defining $R$ can be equivalently expressed in a few ways, as summarized in the following corollary.

\begin{corollary}\label{corollary3.1}
For a clique $G \subseteq P$, an issue $t \in T$, and two thresholds $\mu$ and $\nu$ satisfying $-1 \leq \nu \leq 0 \leq \mu \leq 1$, we have: 
\begin{enumerate}[label = (\arabic*),font = \normalfont]
\setlength{\itemsep}{0pt}
\item $R(G,t) = +1 \Leftrightarrow \mathbb{SA}^{+}(G,t) - \mathbb{SA}^{-}(G,t) \in (\mu,1] \\
\text{\qquad\qquad\quad~~~} \Leftrightarrow \mathbb{SA}^{+}(G,t) \in (\frac{1+\mu}{2},1] \\
\text{\qquad\qquad\quad~~~} \Leftrightarrow \mathbb{SA}^{-}(G,t) \in [0,\frac{1-\mu}{2}) \\
\text{\qquad\qquad\quad~~~} \Leftrightarrow \rho_{t}^{+}(G) - \rho_{t}^{-}(G) \in (\mu,1];$

\item $R(G,t) = 0 \Leftrightarrow \mathbb{SA}^{+}(G,t) - \mathbb{SA}^{-}(G,t) \in [\nu,\mu] \\
\text{\qquad\qquad~~~~} \Leftrightarrow \mathbb{SA}^{+}(G,t) \in [\frac{1+\nu}{2},\frac{1+\mu}{2}] \\
\text{\qquad\qquad~~~~} \Leftrightarrow \mathbb{SA}^{-}(G,t) \in [\frac{1-\mu}{2},\frac{1-\nu}{2}] \\
\text{\qquad\qquad~~~~} \Leftrightarrow \rho_{t}^{+}(G) - \rho_{t}^{-}(G) \in [\nu,\mu];$

\item $R(G,t) = -1 \Leftrightarrow \mathbb{SA}^{+}(G,t) - \mathbb{SA}^{-}(G,t) \in [-1,\nu) \\
\text{\qquad\qquad\quad~~~} \Leftrightarrow \mathbb{SA}^{+}(G,t) \in [0,\frac{1+\nu}{2}) \\
\text{\qquad\qquad\quad~~~} \Leftrightarrow \mathbb{SA}^{-}(G,t) \in (\frac{1-\nu}{2},1] \\
\text{\qquad\qquad\quad~~~} \Leftrightarrow \rho_{t}^{+}(G) - \rho_{t}^{-}(G) \in [-1,\nu)$.
\end{enumerate}
\end{corollary}

\begin{example}\textbf{(Continuing with Example \ref{ex3.1})}\label{ex3.2}
We illustrate the computation of the overall rating of a clique regarding a single issue. 
Taking $\mu = 0.25$ and $\nu = -0.25$, by Definition~\ref{def_R} and Corollary \ref{corollary3.1}, since
$\rho_{t_{1}}^{+}(G)-\rho_{t_{1}}^{-}(G) = \frac{1}{7}-\frac{5}{14} \approx -0.21 \in [-0.25,0.25]$, 
we have $R(G,t_1) = 0$. This indicates that the overall rating of $G$ regarding the issue $t_{1}$ is neutral.
Similarly, for the other issues, we obtain: $R(G,t_2) = +1$, $R(G,t_3) = -1$, $R(G,t_4) = 0$, and $R(G,t_5) = +1$. 
If we use a vector $\bm{\mathrm{R}}_{G} = \big( R(G,t_1), R(G,t_2), \dots, R(G,t_n) \big)$ to express the overall ratings of $X$ about all issues, we have $\bm{\mathrm{R}}_{G} = ( 0,+1,-1,0,+1 )$.
\end{example}

The consistency of a clique $G$ concerning a single issue $t \in T$ can be measured through the distances between individual agents' ratings to the overall rating $R(G,t)$, taking a weighted aggregation.

\begin{definition}
The consistency degree for a clique on a single issue is defined by a function $\mathbb{CM}: 2^{P} \times T \rightarrow [0, 1]$ as: 
\begin{align}\label{eq_CM-i}
\mathbb{CM}(G,t) = 1 - \frac{1}{2} \sum\limits_{p \in G} \big( \theta(p|G) \times |r(p,t) - R(G,t)| \big).
\end{align}
\end{definition}

A weighted aggregation of $\mathbb{CM}$ over individual issues from an issue set $J \subseteq T$ indicates the consistency of $G$ regarding $J$.

\begin{definition}\label{def_CM-J}
The consistency degree for a clique on a non-empty issue set is defined by a function $\mathbb{CM}': 2^{P} \times 2^{T} \rightarrow [0, 1]$ as: 
\begin{align}\label{eq_CM-J}
\mathbb{CM}(G,J) = \sum\limits_{t \in J} \big( \omega(t|J) \times \mathbb{CM}(G,t) \big).
\end{align}
\end{definition}

We will again omit the prime sign in $\mathbb{CM}'$ if the type of parameters is not confusing. {\color{black}By incorporating the weights of both agents and issues, Definition \ref{def_CM-J} presents the calculation of the consistency degree for a clique on a non-empty issue set, denoted as $\mathbb{CM}(G,J)$. The weighting process can be implemented through various established methodologies, including but not limited to: the expert judgment, the analytic hierarchy process, the entropy weight method, and the maximum deviation method.}

One can easily verify that our consistency measures $\mathbb{CM}(G,t)$ and $\mathbb{CM}(G,J)$ cover Xu's measures (i.e. Equations \eqref{eq_CMi-Xu} and \eqref{eq_CMJ-Xu}) as special cases where $\mu = \nu = 0$, and all agents and issues take equal weights. Furthermore, the values of $\mathbb{CM}(G,t)$ can be computed through the positive, neutral, and negative powers of $G$ on $t$, as given in the following theorem. The proof can be found in Appendix A.

\begin{theorem}\label{theorem3.1}
The consistency degree $\mathbb{CM}(G,t)$ of a clique $G \subseteq P$ on an issue $t \in T$ can be computed as:
\begin{align*}
\mathbb{CM}(G,t) 
= \left\{
\begin{array}{lcl}
\rho_{t}^{+}(G) + \frac{1}{2}\rho_{t}^{0}(G),  && \text{if~} R(G,t) = +1, \vspace{2mm} \\
\frac{1}{2}\rho_{t}^{+}(G) + \rho_{t}^{0}(G) + \frac{1}{2}\rho_{t}^{-}(G), && \text{if~} R(G,t) = 0, \vspace{2mm} \\
\rho_{t}^{-}(X) + \frac{1}{2}\rho_{t}^{0}(G),  && \text{if~} R(G,t) = -1.
\end{array}
\right.
\end{align*}
\end{theorem}

This theorem, together with Corollary \ref{corollary3.1} and Proposition \ref{proposition_rho}, helps narrow down the values of $\mathbb{CM}$ and $\mathbb{CM}'$ to the interval $[0.5,1]$.

\begin{theorem}\label{theorem3.2}
For a clique $G \subseteq P$, an issue $t \in T$, and an issue set $J \subseteq T$, we have $\mathbb{CM}(G,t)\in[0.5,1]$ and $\mathbb{CM}(G,J)\in[0.5,1]$.
\end{theorem}

The proof of this theorem can be found in Appendix B.

\begin{example}\textbf{(Continuing with Example \ref{ex3.2})}\label{ex3.3}
This example illustrates the calculation of the consistency degree. Taking the issue $t_{1}$ as an example, since $R(G,t_1) = 0$, by Theorem \ref{theorem3.1}, we have: 
\begin{align*}
\mathbb{CM}(G,t_{1}) = \frac{1}{2}\rho_{t_{1}}^{+}(G) + \rho_{t_{1}}^{0}(G) + \frac{1}{2}\rho_{t_{1}}^{-}(G)
= \frac{1}{2} \times \frac{1}{7} + \frac{1}{2} + \frac{1}{2} \times \frac{5}{14}
= \frac{3}{4}.    
\end{align*}
Similarly, we can compute the following consistency degrees of $G$ with respect to the other issues as: 
\begin{align*}
\mathbb{CM}(G,t_{2}) = \frac{9}{14}, \quad \mathbb{CM}(G,t_{3}) = \frac{9}{14}, \quad 
\mathbb{CM}(G,t_{4}) = \frac{3}{4}, \quad \mathbb{CM}(G,t_{5}) = \frac{5}{7}.    
\end{align*}
{\color{black}The weight vector of all issues is given as $W = (\omega(t_{1}), \omega(t_{2}), \dots, \omega(t_{5})) = (0.3, 0.1, 0.25,  0.15, 0.2)$ based on the expert judgment and domain experience.} For an issue set $J = \{t_{1}, t_{2}, t_{4}\}$, we compute the weight of each issue with respect to $J$ as follows:
\begin{align*}
\omega(t_{1}|J) &= \frac{0.3}{0.3 + 0.1 + 0.15} = \frac{6}{11}, \\
\omega(t_{2}|J) &= \frac{0.1}{0.3 + 0.1 + 0.15} = \frac{2}{11}, \\
\omega(t_{4}|J) &= \frac{0.15}{0.3 + 0.1 + 0.15} = \frac{3}{11}.
\end{align*}
Accordingly, by Equation \eqref{eq_CM-J}, the consistency degree $\mathbb{CM}(G,J)$ of $G$ regarding $J$ is computed as:
\begin{align*}
\mathbb{CM}(G,J) = \frac{6}{11} \times \frac{3}{4} + \frac{2}{11} \times \frac{9}{14} + \frac{3}{11} \times \frac{3}{4} 
= \frac{225}{308}.    
\end{align*}
\end{example}

The consistency degree of a clique $G$ reflects whether the agents' ratings are close or divergent. Therefore, if the consistency degree is large enough, the agents in $G$ can be considered to be allied; if it is small enough, the agents can be considered to be in conflict; otherwise, they are generally neutral to each other.
As a single issue $t \in T$ can be equivalently deemed as a special case of $J = \{t\}$, we will consider a set of issues only in the following discussion.

\begin{definition}
For two thresholds $\beta_{\mathbb{C}}$ and $\alpha_{\mathbb{C}}$ satisfying $0.5 \leq \beta_\mathbb{C} \leq \alpha_\mathbb{C} \leq 1$, the state of a clique $G \subseteq P$ concerning an issue set $J \subseteq T$ is defined as: 
\begin{enumerate}[label = (\arabic*)]
\setlength{\itemsep}{0pt}
\item $G$ is in an \textbf{alliance state} if $\mathbb{CM}(G,J) \geq \alpha_\mathbb{C}$;
\item $G$ is in a \textbf{neutral state} if $\beta_\mathbb{C} < \mathbb{CM}(G,J) < \alpha_\mathbb{C}$;
\item $G$ is in a \textbf{conflict state} if $\mathbb{CM}(G,J) \leq \beta_\mathbb{C}$.
\end{enumerate}
\end{definition}

We used the subscript $\mathbb{C}$ for the two thresholds to distinguish them from those in our approach with non-consistency measures (in Section \ref{4jie}). Considering the consistency measure $\mathbb{CM}$, the three states of $G$ construct a trisection of all issues in $T$.

\begin{definition}
For two thresholds $\beta_{\mathbb{C}}$ and $\alpha_{\mathbb{C}}$ satisfying $0.5 \leq \beta_\mathbb{C} \leq \alpha_\mathbb{C} \leq 1$, the alliance issues $T_{\mathbb{C}}^{=}(G)$, neutral issues $T_{\mathbb{C}}^{\approx}(G)$, and conflict issues $T_{\mathbb{C}}^{\asymp}(G)$ of a clique $G\subseteq P$ based on the consistency measure $\mathbb{CM}$ are defined as: 
\begin{align}
T_{\mathbb{C}}^{=}(G) &= \{t \in T \mid \mathbb{CM}(G,t) \geq \alpha_{\mathbb{C}}\}, \notag\\
T_{\mathbb{C}}^{\approx}(G) &= \{t \in T \mid \beta_{\mathbb{C}} < \mathbb{CM}(G,t) < \alpha_{\mathbb{C}}\},\\
T_{\mathbb{C}}^{\asymp}(G) &= \{t \in T \mid \mathbb{CM}(G,t) \leq \beta_{\mathbb{C}}\}. \notag
\end{align}
\end{definition}

Apparently, $T_{\mathbb{C}}^{=}(G)$, $T_{\mathbb{C}}^{\approx}(G)$, and $T_{\mathbb{C}}^{\asymp}(G)$ form a weak tri-partition of $T$. That is, they satisfy the properties: 
$T_{\mathbb{C}}^{=}(G) \cap T_{\mathbb{C}}^{\approx}(G) = \emptyset$, 
$T_{\mathbb{C}}^{=}(G) \cap T_{\mathbb{C}}^{\asymp}(G) = \emptyset$, 
$T_{\mathbb{C}}^{\asymp}(G) \cap T_{\mathbb{C}}^{\approx}(G) = \emptyset$, 
and $T_{\mathbb{C}}^{=}(G) \cup T_{\mathbb{C}}^{\approx}(G) \cup T_{\mathbb{C}}^{\asymp}(G) = T$.

\subsection{Selecting optimal feasible strategies with consistency measures}\label{3.3jie}\hspace{0.2in}
We present the methods for identifying feasible strategies and optimal feasible strategies based on the consistency measures. The corresponding $L$-order strategies will be also discussed. We first formally define the strategies and $L$-order strategies.

\begin{definition}\label{def_ST}
Any non-empty subset of issues $J \subseteq T$ is called a strategy. The set of all strategies is represented as $ST = \{J \subseteq T \mid J \neq \emptyset\}$.
\end{definition}

Evidently, the number of strategies is given by $\#(ST) = 2^{\#(T)} - 1 = 2^{n} - 1$.

\begin{definition}\label{def_ST-L}
A non-empty subset of issues $J \in ST$ is called an $L$-order strategy where $1 \leq L \leq n$ if $\#(J) = L$. Moreover, we denote the set of all $L$-order strategies as $ST_{L}=\{J\in ST \mid \#(J)=L\}$.
\end{definition}

The set $ST_{L}$ consists of all non-empty subsets of $T$ with a cardinality of $L$. Thus, we have $\#(ST_{L})=C_{n}^{L}=\frac{n!}{L!(n-L)!}$.

We interpret and evaluate the feasibility of a strategy in terms of consistency measures. More specifically, a feasible strategy should have a large-enough consistency degree, evaluated concerning a large-enough clique of agents.

\begin{definition}\label{def_FS-C}
A strategy $J \in ST$ is called a feasible strategy for a clique of agents $G \subseteq T$ regarding the consistency measure $\mathbb{CM}'$ if: 
\begin{align}
\frac{\#(G)}{\#(P)} \geq \gamma_P  \wedge \mathbb{CM}(G,J) \geq \lambda,
\end{align}
where $\gamma_P \in [0, 1]$ and $\lambda \in [0.5, 1]$ are two given thresholds. The set of all feasible strategies for the clique $G$ obtained by the consistency measure $\mathbb{CM}'$ is denoted by $FS^{\mathbb{C}}(G)$.
\end{definition}

An $L$-order feasible strategy is similarly defined with an additional requirement of exactly $L$ issues.

\begin{definition}\label{def_FS-LC}
An $L$-order strategy $J \in ST_{L}$ is called an $L$-order feasible strategy for a clique of agents $G \subseteq P$ with respect to a consistency measure $\mathbb{CM}'$ if: 
\begin{align}
\frac{\#(G)}{\#(P)} \geq \gamma_P  \wedge \mathbb{CM}(G,J) \geq \lambda,
\end{align}
where $\gamma_P \in [0, 1]$ and $\lambda \in [0.5, 1]$ are two given thresholds. The set of all $L$-order feasible strategies for the clique $G$ obtained by the consistency measure $\mathbb{CM}'$ is denoted by $FS^{\mathbb{C}}_{L}(G)$.
\end{definition}

An optimal feasible strategy takes the maximum consistency degree and should involve a large-enough number of issues.

\begin{definition}\label{def_FS-Copt} 
An optimal feasible strategy for a clique $G \subseteq P$ concerning the consistency measure $\mathbb{CM}'$ is: 
\begin{align}\label{eq_FS-Copt}
\widehat{J} \in \{ \argmax\limits_{J\in FS^{\mathbb{C}}(G) \wedge \frac{\#(J)}{\#(T)} \geq \gamma_T} \mathbb{CM}(G,J) \},
\end{align}
where $\gamma_T \in [0,1]$ is a given threshold. The set of all optimal feasible strategies for the clique $G$ obtained by consistency measure $\mathbb{CM}'$ is denoted as $\widehat{FS}^{\mathbb{C}}(G)$.
\end{definition}

Equation \eqref{eq_FS-Copt} expresses two conditions: (1) An optimal feasible strategy $\widehat{J}\in \widehat{FS}^{\mathbb{C}}(G)$ must be relatively large enough compared to $T$, and (2) it has the maximum consistency degree among all feasible strategies. Similarly, we also obtain an $L$-order optimal feasible strategy.

\begin{definition}\label{def_FS-LCopt} 
An $L$-order optimal feasible strategy for a clique $G \subseteq P$ with respect to the consistency measure $\mathbb{CM}'$ is: 
\begin{align}\label{eq_FS-LCopt}
\widehat{J}_{L} \in \{ \argmax\limits_{J\in FS_{L}^{\mathbb{C}}(G) \wedge \frac{\#(J)}{\#(T)} \geq \gamma_T} \mathbb{CM}(G,J) \},
\end{align}
where $\gamma_T \in [0,1]$ is a given threshold. The set of all $L$-order optimal feasible strategies for the clique $G$ obtained by the consistency measure $\mathbb{CM}'$ is denoted as $\widehat{FS}_{L}^{\mathbb{C}}(G)$.
\end{definition}

By the above definitions, we present Algorithm \ref{algorithm1} for identifying the sets of feasible strategies and $L$-order feasible strategies, and Algorithm \ref{algorithm2} for determining the sets of optimal feasible strategies and $L$-order optimal feasible strategies for a given clique of agents. 
{\color{black}
In Algorithm \ref{algorithm1}, the time complexity of Lines 2-11 is $O(\#(G)\times\#(T))$, the time complexity of Lines 12-13 is $O(2^{\#(T)})$, and the time complexity of Lines 14-22 is $O(2^{\#(T)-1})$. Thus, the total time complexity of Algorithm \ref{algorithm1} is $O(\max\{\#(G)\times\#(T),2^{\#(T)}\})$ and the total space complexity of Algorithm \ref{algorithm1} is $O(\#(T)\times2^{\#(T)})$.
In Algorithm \ref{algorithm2}, the time complexity of Lines 2-8 is $O(\#(FS^{\mathbb{C}}(G)))$ and the time complexity of Lines 9-15 is $O(\#(FS_{L}^{\mathbb{C}}(G)))$. Thus, the total time complexity of Algorithm \ref{algorithm1} is $O(\#(FS^{\mathbb{C}}(G)))$ and the total space complexity of Algorithm \ref{algorithm1} is $O(\#(T)\times\#(FS^{\mathbb{C}}(G)))$.
}

\begin{algorithm}[!ht]
	\setstretch{1.25}
	\caption{The algorithm of selecting the sets of feasible strategies $FS^{\mathbb{C}}(G)$ and  $L$-order feasible strategies $FS_{L}^{\mathbb{C}}(G)$ for a clique $G$.}
	\label{algorithm1}
	\SetKwData{Left}{left}\SetKwData{This}{this}\SetKwData{Up}{up}
	\SetKwFunction{Union}{Union}\SetKwFunction{FindCompress}{FindCompress}
	\SetKwInOut{Input}{Input}\SetKwInOut{Output}{Output}
	
	\Input{A three-valued situation table $TS = (P, T, V, r)$; \\
		The target clique $G \subseteq P$ satisfying $\frac{\#(G)}{\#(P)} \geq \gamma_{P}$; \\
		The weight vector of agents $\Theta=(\theta(p_{1}), \theta(p_{2}), \dots, \theta(p_{m}))$; \\
		The weight vector of issues $W=(\omega(t_{1}), \omega(t_{2}), \dots, \omega(t_{n}))$; \\
		Parameters $\gamma_{P} \in [0,1]$, $\lambda \in [0.5,1]$, and $L \in \{1,2,...,n\}$. }
	\Output{The sets of feasible strategies $FS^{\mathbb{C}}(G)$ and 
		  $L$-order feasible strategies $FS_{L}^{\mathbb{C}}(G)$ of $G$.}
	
	\Begin{		
		\For{each issue $t \in T$}{
            \For{each agent $p \in G$ }{
			    {\textbf{Compute} $\mathbb{SA}^{+}(p,t) = 1-\frac{|r(p,t)-r(p^{+},t)|}{2}$ by Equation~\eqref{eq_SA+ai}.}	
			
			    {\textbf{Compute} $\mathbb{SA}^{-}(p,t) = 1-\frac{|r(p,t)-r(p^{-},t)|}{2}$ by Equation~\eqref{eq_SA-ai}.}
            }
			
			{\textbf{Compute} 
			$\mathbb{SA}^{+}(G,t) = \sum\limits_{p \in G} \big( \theta(p|G) \times \mathbb{SA}^{+}(p,t) \big)$ by Equation~\eqref{eq_SA+Xi}.}
			
			{\textbf{Compute} 
			$\mathbb{SA}^{-}(G,t) = \sum\limits_{p \in G} \big( \theta(p|G) \times \mathbb{SA}^{-}(p,t) \big)$ by Equation~\eqref{eq_SA-Xi}.}
			
			{\textbf{Compute} $R(G,t)$ by Equation \eqref{eq_At}.}
			
			{\textbf{Compute} 
			$\mathbb{CM}(G,t) = 1 - \frac{1}{2} \sum\limits_{p \in G} \big( \theta(p|G) \times |r(p, t) - R(G,t)| \big)$ by Equation \eqref{eq_CM-i}.}
        }
		
		\textbf{Initialize} $FS^{\mathbb{C}}(G) = \emptyset$ and $FS_{L}^{\mathbb{C}}(G) = \emptyset$.
		
		\textbf{Create} the set of strategies $ST = \{ J \subseteq T \mid J \neq \emptyset \}$.
		
		\For{each strategy $J\in ST$}{
			{\textbf{Compute} $\mathbb{CM}(G,J) = \sum\limits_{t \in J} \big( \omega(t|J) \times \mathbb{CM}(G,t)\big)$ by Equation \eqref{eq_CM-J}}. 
			
			\If{$\frac{\#(G)}{\#(P)} \geq \gamma_{P}$ and $\mathbb{CM}(G,J) \geq \lambda$}{
			$FS^{\mathbb{C}}(G) = FS^{\mathbb{C}}(G) \cup \{J\}$ by Definition~\ref{def_FS-C}.
				
				\If{$\#(J)=L$}{
				$FS_{L}^{\mathbb{C}}(G) = FS_{L}^{\mathbb{C}}(G) \cup \{J\}$ by Definition~\ref{def_FS-LC}.
				}
			}
		}

	}
	\textbf{Return:} $FS^{\mathbb{C}}(G)$ and $FS_{L}^{\mathbb{C}}(G)$.
\end{algorithm}

\begin{algorithm}[!ht]
\setstretch{1.25}
\caption{The algorithm of selecting the sets of optimal feasible strategies $\widehat{FS}^{\mathbb{C}}(G)$ and $L$-order optimal feasible strategies $\widehat{FS}_{L}^{\mathbb{C}}(G)$ for a clique $G$.}
\label{algorithm2}
\SetKwData{Left}{left}\SetKwData{This}{this}\SetKwData{Up}{up}
\SetKwFunction{Union}{Union}\SetKwFunction{FindCompress}{FindCompress}
\SetKwInOut{Input}{Input}\SetKwInOut{Output}{Output}

\Input{The set of feasible strategies $FS^{\mathbb{C}}(G)$ of $G$; \\
    The consistency degree of each strategy in $FS^{\mathbb{C}}(G)$; \\
	The set of $L$-order feasible strategies $FS_{L}^{\mathbb{C}}(G)$ of $G$; \\
	The consistency degree of each $L$-order strategy in $FS_{L}^{\mathbb{C}}(G)$; \\
    Parameter $\gamma_T \in [0,1]$.}
\Output{The set of optimal feasible strategies $\widehat{FS}^{\mathbb{C}}(G)$ of $G$; \\
	The set of $L$-order optimal feasible strategies $\widehat{FS}_{L}^{\mathbb{C}}(G)$ of $G$.}

\Begin{	
\textbf{Compute} the maximum consistency degree 
$\mathbb{CM}^{max} = \max\limits_{J\in FS^{\mathbb{C}}(G)} \mathbb{CM}(G,J)$. \\

\textbf{Initialize} $\widehat{FS}^{\mathbb{C}}(G) = \emptyset$.

\For{each feasible strategy $J \in FS^{\mathbb{C}}(G)$}{
	\If{$\frac{\#(J)}{\#(T)} \geq \gamma_T$ and $\mathbb{CM}(G,J) = \mathbb{CM}^{max}$}{
	$\widehat{FS}^{\mathbb{C}}(G) = \widehat{FS}^{\mathbb{C}}(G) \cup \{J\}$ by Definition \ref{def_FS-Copt}.
	}
}

\textbf{Compute} the maximum consistency degree 
$\mathbb{CM}^{max}_{L} = \max\limits_{J_{L}\in FS_{L}^{\mathbb{C}}(G)}\mathbb{CM}(G,J_{L})$. \\

\textbf{Initialize} $\widehat{FS}_{L}^{\mathbb{C}}(G)=\emptyset$.

\For{each $L$-order feasible strategy $J_{L}\in FS_{L}^{\mathbb{C}}(G)$}{
	\If{$\frac{\#(J_{L})}{\#(T)} \geq \gamma_T$ and $\mathbb{CM}(G,J_{L}) = \mathbb{CM}^{max}_{L}$}{
	$\widehat{FS}_{L}^{\mathbb{C}}(G) = \widehat{FS}_{L}^{\mathbb{C}}(G) \cup \{J_{L}\}$ by Definition \ref{def_FS-LCopt}.
	}
}

}
\textbf{Return:} $\widehat{FS}^{\mathbb{C}}(G)$ and $\widehat{FS}_{L}^{\mathbb{C}}(G)$.
\end{algorithm}

\begin{example}\textbf{(Continuing with Example \ref{ex3.3})}\label{ex3.4}
This example illustrates our above discussion regarding strategies based on consistency measures. 
Assume we have $\gamma_{P}=0.5$. Since $0.5\times\#(P)=3$, the clique $G$ should include at least three agents. 
We consider the clique $G = \{p_{1}, p_{3}, p_{4}, p_{6}\}$.
Firstly, by Definition~\ref{def_ST}, we identify the set of strategies $ST$ and calculate the consistency degrees for each strategy, as presented in Table \ref{tab_ST-C}. 
Secondly, assuming $L=3$, by Definition \ref{def_ST-L}, we can easily identify a set of ten $3$-order strategies $ST_{3}$. 
Then, assuming $\lambda = 0.73$, by Definition \ref{def_FS-C}, we obtain the set of feasible strategies for the clique $G$ as: 
\begin{align*}
FS^{\mathbb{C}}(G) = \{ \{t_1\}, \{t_4\}, \{t_1, t_4\}, \{t_1, t_5\}, \{t_1, t_2, t_4\}, \{t_1, t_4, t_5\} \}.
\end{align*}
By Definition \ref{def_FS-LC}, we obtain the set of $3$-order feasible strategies for the clique $G$ as:
\begin{align*}
FS_{3}^{\mathbb{C}}(G) = \{ \{t_1, t_2, t_4\}, \{t_1, t_4, t_5\} \}.
\end{align*}
In Table \ref{tab_ST-C}, all feasible strategies are shaded in green, and all $3$-order feasible strategies are highlighted in blue.
Finally, with $\gamma_T = 0.5$, by Definitions \ref{def_FS-Copt} and \ref{def_FS-LCopt}, we determine the sets of optimal feasible strategy and $3$-order optimal feasible strategy for the clique $G$: 
\begin{align*}
\widehat{FS}^{\mathbb{C}}(G) = \widehat{FS}_{3}^{\mathbb{C}}(G) = \{ \{t_1, t_4, t_5\} \},  
\end{align*}
which is underlined in Table \ref{tab_ST-C}.

\begin{table}[!ht]
	\renewcommand{\arraystretch}{1.3}
	\begin{center}
		\caption{Strategies and their consistency degrees with respect to $G$.}\label{tab_ST-C}
		\setlength{\tabcolsep}{5.3mm}
		\begin{tabular}{lc|lc|lc}
			\hline
			$J$  & $\mathbb{CM}(G,J)$ & $J$ & $\mathbb{CM}(G,J)$ & $J$ & $\mathbb{CM}(G,J)$ \\
			\hline
			\cellcolor{green!20}$\{t_{1}\}$ & \cellcolor{green!20}$\frac{3}{4}$ 
			& $\{t_{2},t_{5}\}$ & $\frac{29}{42}$   
			& $\{t_{2},t_{3},t_{5}\}$ & $\frac{103}{154}$ \\
			
			$\{t_{2}\}$ & $\frac{9}{14}$ 
			& $\{t_{3},t_{4}\}$ & $\frac{153}{224}$ 
			& $\{t_{2},t_{4},t_{5}\}$  & $\frac{179}{252}$ \\
			
			$\{t_{3}\}$ &$\frac{9}{14}$ 
			& $\{t_{3},t_{5}\}$ & $\frac{85}{126}$  
			& $\{t_{3},t_{4},t_{5}\}$ & $\frac{233}{336}$ \\
			
			\cellcolor{green!20}$\{t_{4}\}$ & \cellcolor{green!20}$\frac{3}{4}$ 
			& $\{t_{4},t_{5}\}$ & $\frac{143}{196}$ 
			& $\{t_{1},t_{2},t_{3},t_{4}\}$ & $\frac{45}{64}$ \\
			
			$\{t_{5}\}$ &$\frac{5}{7}$  
			& $\{t_{1},t_{2},t_{3}\}$ & $\frac{9}{13}$ 
			& $\{t_{1},t_{2},t_{3},t_{5}\}$ & $\frac{83}{119}$ \\
			
			$\{t_{1},t_{2}\}$ &$\frac{81}{112}$ 
			& \color{blue}\cellcolor{green!20}$\{t_{1},t_{2},t_{4}\}$ 
			& \color{blue}\cellcolor{green!20}$\frac{225}{308}$ 
			& $\{t_{1},t_{2},t_{4},t_{5}\}$ & $\frac{61}{84}$ \\
			
			$\{t_{1},t_{3}\}$ &$\frac{54}{77}$ 
			& $\{t_{1},t_{2},t_{5}\}$ & $\frac{121}{168}$ 
			& $\{t_{1},t_{3},t_{4},t_{5}\}$ & $\frac{359}{504}$ \\
			
			\cellcolor{green!20}$\{t_{1},t_{4}\}$ & \cellcolor{green!20}$\frac{3}{4}$ 
			& $\{t_{1},t_{3},t_{4}\}$ & $\frac{279}{392}$ 
			& $\{t_{2},t_{3},t_{4},t_{5}\}$ & $\frac{269}{392}$ \\
			
			\cellcolor{green!20}$\{t_{1},t_{5}\}$ & \cellcolor{green!20}$\frac{103}{140}$ 
			& $\{t_{1},t_{3},t_{5}\}$ & $\frac{74}{105}$ 
			& $\{t_{1},t_{2},t_{3},t_{4},t_{5}\}$ & $\frac{79}{112}$\\
			
			$\{t_{2},t_{3}\}$ & $\frac{9}{14}$ 
			& \color{blue}\cellcolor{green!20}\underline{$\{t_{1},t_{4},t_{5}\}$} 
			& \color{blue}\cellcolor{green!20}\underline{$\frac{269}{364}$} \\
			
			$\{t_{2},t_{4}\}$ & $\frac{99}{140}$ 
			& $\{t_{2},t_{3},t_{4}\}$ & $\frac{27}{40}$ \\
			\hline
		\end{tabular}
	\end{center}
\end{table}
\end{example}

\section{Constructing feasible strategies with non-consistency measures}\label{4jie}\hspace{0.2in}
In this section, we take a different perspective from Section \ref{3jie} and propose non-consistency measures to select feasible strategies.

\subsection{Non-consistency measures incorporating weights of agents and issues}\label{4.1jie}\hspace{0.2in}
First of all, we introduce a novel conflict measure on agents that incorporates their weights.

\begin{definition}\label{def_CA-i}
The conflict degree between two agents on an issue $t\in T$ is defined by a conflict measure $\mathbb{CA}_{t}: P \times P \rightarrow [0,1]$ as: 
\begin{align}\label{eq_CA-i}
\mathbb{CA}_{t}(p,q) = \frac{\left| r(p,t)-r(q,t) \right| \times (1 - \left| \theta(p|pq)-\theta(q|pq) \right|}{2},
\end{align}
where $\theta(p|pq)$ and $\theta(q|pq)$ are shorthand notations of $\theta(p|\{p,q\})$ and $\theta(q|\{p,q\})$, that is, the weights of agents $p$ and $q$ regarding the clique $\{p,q\}$ calculated as: 
\begin{align}\label{eq_theta-ab}
\theta(p|pq) = \frac{\theta(p)}{\theta(p) + \theta(q)}, \quad
\theta(q|pq) = \frac{\theta(q)}{\theta(p) + \theta(q)}. 
\end{align}
\end{definition}

As $\theta(p|pq)+\theta(q|pq)=1$, one may alternatively compute $\mathbb{CA}_{t}(p,q)$ as given in the following theorem. The proof is straightforward and thus, omitted.

\begin{theorem}\label{theorem4.1}
The conflict degree $\mathbb{CA}_{t}(p,q)$ between agents $p$ and $q$ on an issue $t \in T$ can be computed as:
\begin{align*}
\mathbb{CA}_{t}(p,q) 
&= \left| r(p,t)-r(q,t) \right| \times \min\{\theta(p|pq),\theta(q|pq)\} \\
&= \left\{
\begin{array}{lcl}
0,   && \text{if~} r(p,t)=r(q,t), \vspace{2mm} \\
\min\{\theta(p|pq),\theta(q|pq)\},   && \text{if~} \big( r(p,t) \times r(p,t) = 0 \big) \wedge \big( r(p,t) \neq r(p,t) \big), \vspace{2mm} \\
2 \times \min\{\theta(p|pq),\theta(q|pq)\},  && \text{if~} r(p,t) \times r(q,t) = -1. 
\end{array}
\right.
\end{align*}
\end{theorem}

Intuitively, the opinion of an agent with a larger weight is more influential. Therefore, when there is a significant imbalance in the weights of two agents, their conflict degree may remain relatively low even if their opinions are opposite. In an extreme case where $\theta(p|pq) = 1$ and $\theta(q|pq) = 0$, the agent $p$ holds full power, while the agent $q$ has none. Even if they have opposing views on a given issue, their conflict degree is $0$. Moreover, unequal weights of agents imply that one agent has a more influential opinion than the other, which will reduce their conflict compared with the case of equal weights, or equivalently, the unweighted case. For example, with equal weights $\theta(p|pq) = \theta(q|pq) = 0.5$, two opposite ratings $r(p,t) = +1$ and $r(q,t) = -1$ lead to a conflict degree of $\mathbb{CA}_{t}(p,q) = 1$. In contrast with unequal weights $\theta(p|pq) = 0.7$, $\theta(q|pq) = 0.3$, and the same ratings, their conflict degree decreases to $\mathbb{CA}_{t}(p,q) = 0.6$. Furthermore, with more imbalanced weights  $\theta(p|pq) = 0.9$ and $\theta(q|pq) = 0.1$, their conflict degree further decreases to $\mathbb{CA}_{t}(p,q) = 0.2$.

In a special case where all agents are assigned the same weight, that is, $\theta(p|pq) = \theta(q|pq) = 0.5$. The conflict degree between agents $p$ and $q$ concerning an issue $t \in T$ is computed as: 
\begin{align}
\mathbb{CA}_{t}(p,q) = \frac{|r(p,t) - r(q,t)|}{2} = \left\{
\begin{array}{lcl}
0,   && \text{if~~} r(p,t) = r(q,t), \vspace{2mm} \\
0.5,   && \text{if~~} \big( r(p,t) \times r(q,t) = 0 \big) \wedge \big( r(p,t) \neq r(q,t) \big), \vspace{2mm} \\
1,  && \text{if~~} r(p,t) \times r(q,t) = -1,
\end{array}
\right.
\end{align}
which is consistent with the distance function used in the existing works \cite{Yao2019}.

Conflict measure evaluates the agent relationships and thus, is commonly used to define the three typical agent relations of alliance, conflict, and neutrality between two agents with respect to a given issue.

\begin{definition}
For two thresholds $\beta_{i}$ and $\alpha_{i}$ satisfying $0 \leq \beta_{t} \leq \alpha_{t} \leq 1$, the alliance relation $R_{t}^{=}$, neutrality relation $R_{t}^{\approx}$, and conflict relation $R_{t}^{\asymp}$ on agents regarding an issue $t \in T$ are defined as:
\begin{align}
R_{t}^{=} &= \{(p,q) \in P \times P \mid \mathbb{CA}_{t}(p,q) \leq \beta_{t}\}, \notag \\
R_{t}^{\approx} &= \{(p,q) \in P \times P \mid \beta_{t} < \mathbb{CA}_{t}(p,q) < \alpha_{t}\},\\
R_{t}^{\asymp} &= \{(p,q) \in P \times P \mid \mathbb{CA}_{t}(p,q) \geq \alpha_{t}\}. \notag
\end{align}
\end{definition}

The three relations form a trisection of all agent pairs in $P \times P$, that is, we have
$R_{t}^{=} \cap R_{t}^{\approx} = \emptyset,  R_{t}^{=} \cap R_{t}^{\asymp} = \emptyset,  R_{t}^{\approx} \cap R_{t}^{\asymp} = \emptyset,  \text{ and } R_{t}^{=} \cup R_{t}^{\approx} \cup R_{t}^{\asymp} = P \times P$.
The above discussion can be generalized to multiple issues.

\begin{definition}\label{def_CA-J}
The conflict degree between two agents on a non-empty issue set $J\subseteq T$ is defined by a conflict measure $\mathbb{CA}_{J} : P \times P \to [0, 1]$ as: 
\begin{align}\label{eq_CA-J}
\mathbb{CA}_{J}(p,q) = \sum_{t \in J} \big( \omega(t|J) \times \mathbb{CA}_{t}(p,q) \big).
\end{align}
\end{definition}

Accordingly, one may easily generalize the agent relations regarding a single issue to those concerning a non-empty issue set.

\begin{definition}
For two thresholds $\beta_{J}$ and $\alpha_{J}$ satisfying $0 \leq \beta_{J} \leq \alpha_{J} \leq 1$, the alliance relation $R_{J}^{=}$, neutrality relation $R_{J}^{\approx}$, and conflict relation $R_{J}^{\asymp}$ on agents concerning a non-empty issue set $J \subseteq T$ are defined as: 
\begin{align}
R_{J}^{=} &= \{ (p,q) \in P \times P \mid \mathbb{CA}_{J}(p,q) \leq \beta_{J} \}, \notag\\
R_{J}^{\approx} &= \{ (p,q) \in P \times P \mid \beta_{J} < \mathbb{CA}_{J}(p,q) < \alpha_{J} \}, \\
R_{J}^{\asymp} &= \{ (p,q) \in P \times P \mid \mathbb{CA}_{J}(p,q) \geq \alpha_{J} \}.\notag
\end{align}
\end{definition}

The three relations again form a trisection of $P\times P$, that is, we have: $R_{J}^{=}\cap R_{J}^{\approx}=\emptyset$, $R_{J}^{=}\cap R_{J}^{\asymp}=\emptyset$, $R_{J}^{\approx}\cap R_{J}^{\asymp}=\emptyset$, and $R_{J}^{=}\cup R_{J}^{\approx}\cup R_{J}^{\asymp}=P \times P$.

One can easily obtain a few properties of $\mathbb{CA}_{t}$ and $\mathbb{CA}_{J}$ as given in the following proposition. The proof is straightforward from Definitions \ref{def_CA-i} and \ref{def_CA-J} and thus, is omitted.

\begin{proposition}\label{proposition4.1}
The following properties hold for an issue $t \in T$, a non-empty subset of issues $J \subseteq T$, and two agents $p, q \in P$: 
\begin{enumerate}[label = (\arabic*), font = \normalfont]
\setlength{\itemsep}{0pt}
\item Symmetry$:$ \\
$\mathbb{CA}_{t}(p,q) = \mathbb{CA}_{t}(q,p);$ \\
$\mathbb{CA}_{J}(p,q) = \mathbb{CA}_{J}(q,p)$.

\item Boundary conditions for $\mathbb{CA}_{t}:$ \\
$\mathbb{CA}_{t}(p,q) = 0 \Leftrightarrow r(p,t) = r(q,t) \vee \theta(p|pq) = 0 \vee \theta(q|pq) = 0;$ \\ 
$\mathbb{CA}_{t}(p,q) = 1 \Leftrightarrow r(p,t) \times r(q,t) = -1 \wedge \theta(p|pq) = \theta(q|pq) = 0.5$.

\item Boundary conditions for $\mathbb{CA}_{J}:$ \\
$\mathbb{CA}_{J}(p,q) = 0 \Leftrightarrow \forall t \in J, \mathbb{CA}_{t}(p,q) = 0;$ \\
$\mathbb{CA}_{J}(p,q) = 1 \Leftrightarrow \forall t \in J, \mathbb{CA}_{t}(p,q) = 1$.
\end{enumerate}
\end{proposition}

\begin{example}\textbf{(Continuing with Example \ref{ex3.4})}\label{ex4.1}
{\color{black}We adopt the same weight vector of agents as specified in Example \ref{ex3.1}, namely $\Theta = (0.25, 0.18, 0.1, 0.15, 0.12, 0.2)$.} We take $p_{1}$ and $p_{2}$ to illustrate the calculation of conflict degrees. 
Using Equation \eqref{eq_theta-ab}, we compute their relative weights as:
\begin{align*}
\theta(p_{1}|p_{1}p_{2}) = \frac{0.25}{0.25 + 0.18} = \frac{25}{43} \quad \text{and} \quad 
\theta(p_{2}|p_{1}p_{2}) = \frac{0.18}{0.25 + 0.18} = \frac{18}{43}.  
\end{align*}
Following Theorem \ref{theorem4.1}, since $r(p_{1},t_{1}) \times r(p_{2},t_{1}) = -1$, the conflict degree between $p_{1}$ and $p_{2}$ regarding the issue $t_{1}$ is:
\begin{align*}
\mathbb{CA}_{t_{1}}(p_{1},p_{2}) = 2 \times \min\{ \theta(p_{1}|p_{1}p_{2}), \theta(p_{2}|p_{1}p_{2}) \} = \frac{36}{43}. 
\end{align*}
Table \ref{tab_CA_i1} gives the conflict degrees between two agents regarding the issue $t_{1}$.

\begin{table}[!ht]
\renewcommand{\arraystretch}{1.3}
\begin{center}
\caption{Conflict degrees between two agents regarding the issue $t_{1}$.}\label{tab_CA_i1}
\setlength{\tabcolsep}{9.6mm}
\begin{tabular}{cccccccc}
\hline
& $p_{1}$ & $p_{2}$ & $p_{3}$ & $p_{4}$ & $p_{5}$ & $p_{6}$\\
\hline
$p_{1}$ & $0$ & $\frac{36}{43}$ & $\frac{4}{7}$ & $\frac{3}{8}$ & $\frac{24}{37}$ & $\frac{4}{9}$ \\
$p_{2}$ & $\frac{36}{43}$ & $0$ & $0$ & $\frac{5}{11}$ & $0$ & $\frac{9}{19}$ \\
$p_{3}$ & $\frac{4}{7}$ & $0$ & $0$ & $\frac{2}{5}$ & $0$ & $\frac{1}{3}$ \\
$p_{4}$ & $\frac{3}{8}$ & $\frac{5}{11}$ & $\frac{2}{5}$ & $0$ & $\frac{4}{9}$ & $0$ \\
$p_{5}$ & $\frac{24}{37}$ & $0$ & $0$ & $\frac{4}{9}$ & $0$ & $\frac{3}{8}$ \\
$p_{6}$ & $\frac{4}{9}$ & $\frac{9}{19}$ & $\frac{1}{3}$ & $0$ & $\frac{3}{8}$ & $0$ \\
\hline
\end{tabular}
\end{center}
\end{table}
\end{example}

The conflict degree between agents in a clique reflects their divergent attitudes towards an issue and thus, can be synthesized to measure the overall non-consistency in the clique.

\begin{definition}\label{def_NM-i}
The non-consistency degree for a clique on a single issue is defined by a function $\mathbb{NM}: 2^{P} \times T \rightarrow [0,1]$ as: 
\begin{align}\label{eq_NM-i}
\mathbb{NM}(G,t) &= \frac{\sum\limits_{(p,q) \in G \times G}\mathbb{CA}_{t}(p,q)}{\big( \#(G) \big)^{2}}.
\end{align}
\end{definition}

The non-consistency regarding a set of issues takes a weighted average of those concerning individual issues in the issue set.

\begin{definition}\label{def_NM-J}
The non-consistency degree for a clique on a non-empty issue set is defined by a function $\mathbb{NM}': 2^{P} \times 2^{T} \rightarrow [0,1]$ as: 
\begin{align}\label{eq_NM-J}
\mathbb{NM}(G,J) &= \sum\limits_{t \in J}\big( \omega(t|J) \times \mathbb{NM}(G,t) \big).
\end{align}
\end{definition}

The prime sign will be omitted without causing confusion. {\color{black}Definition \ref{def_NM-J} gives the calculation of the non-consistency degree for a clique on a non-empty issue set, denoted as $\mathbb{NM}(G,J)$, which also takes into account the weights of both agents and issues. These weights are determined using the similar established methods to Section \ref{3jie}.} By Definitions \ref{def_CA-J}, \ref{def_NM-i}, and \ref{def_NM-J}, one can measure $\mathbb{NM}(G,J)$ directly through the conflict degrees.

\begin{theorem}
The non-consistency degree $\mathbb{NM}(G,J)$ of a clique $G \subseteq P$ with respect to a non-empty issue set $J \subseteq T$ can be computed as:
\begin{align*}
\mathbb{NM}(G,J) = \frac{\sum\limits_{(p,q)\in G \times G}\mathbb{CA}_{J}(p,q)}{\big( \#(G) \big)^{2}}.
\end{align*}
\end{theorem}

{\color{black}
\begin{proof}
According to Equations \eqref{eq_NM-i} and \eqref{eq_NM-J}, we have:
\begin{align*}
\mathbb{NM}(G,J) &= \frac{\sum\limits_{t \in J} \big( \omega(t|J) \times \sum\limits_{(p,q)\in G\times G}\mathbb{CA}_{t}(p,q) \big)}{\big( \#(G) \big)^{2}} \\
&= \frac{\sum\limits_{(p,q)\in G\times G} \sum\limits_{t \in J} \big( \omega(t|J) \times \mathbb{CA}_{t}(p,q) \big)}
{\big( \#(G) \big)^{2}} \\
&= \frac{\sum\limits_{(p,q)\in G\times G} \mathbb{CA}_{J}(p,q)}{\big( \#(G) \big)^{2}}.
\end{align*}
Thus, the theorem is proved.
\end{proof}}

\begin{example}\textbf{(Continuing with Example~\ref{ex4.1})}\label{ex4.2}
{\color{black}We adopt the same weight vector of issues as specified in Example \ref{ex3.3}, namely $W = (0.3, 0.1, 0.25,  0.15, 0.2)$.} This example illustrates the calculation of non-consistency degrees. 
With Table \ref{tab_CA_i1}, we calculate the non-consistency degree of $G = \{p_{1},p_{3},p_{4},p_{6}\}$ on the issue $i_{1}$ as: 
\begin{align*}
\mathbb{NM}(G,t_{1}) &= \frac{1}{4^2} \times \left( 0 + \frac{4}{7} + \frac{3}{8} + \frac{4}{9} + \frac{4}{7} + 0 + \frac{2}{5} + \frac{1}{3} + \frac{3}{8} + \frac{2}{5} + 0 + 0 + \frac{4}{9} + \frac{1}{3} + 0 + 0 \right) \\
&\approx 0.2655.
\end{align*}
Similarly, we can get the non-consistency degrees regarding the other issues:
\begin{align*}
\mathbb{NM}(G,t_{2}) = 0.3557, \quad \mathbb{NM}(G,t_{3}) = 0.2763, \quad
\mathbb{NM}(G,t_{4}) = 0.2655, \quad \mathbb{NM}(G,t_{5}) = 0.3283.
\end{align*}
Then, by Definition~\ref{def_NM-J}, the non-consistency degree of the clique $G$ regarding $J = \{t_{1},t_{2},t_{4}\}$ is calculated as:
\begin{align*}
\mathbb{NM}(X,J) = \frac{6}{11} \times 0.2655 + \frac{2}{11} \times 0.3557 + \frac{3}{11} \times 0.2655 = 0.2819.
\end{align*}
\end{example}

The state of a clique can be determined by the non-consistency measure. We'll again consider issue sets as a general case.

\begin{definition}
For two thresholds $\beta_{\mathbb{N}}$ and $\alpha_{\mathbb{N}}$ satisfying $0 \leq \beta_{\mathbb{N}} \leq \alpha_{\mathbb{N}} \leq 1$, the state of a clique $G \subseteq P$ concerning an issue set $J \subseteq T$ is defined as:
\begin{enumerate}[label = (\arabic*)]
\setlength{\itemsep}{0pt}
\item $G$ is in an \textbf{alliance state} if $\mathbb{NM}(G,J) \leq \beta_{\mathbb{N}}$;
\item $G$ is in a \textbf{neutral state} if $\beta_{\mathbb{N}} < \mathbb{NM}(G,J) < \alpha_{\mathbb{N}}$;
\item $G$ is in a \textbf{conflict state} if $\mathbb{NM}(G,J) \geq \alpha_{\mathbb{N}}$.
\end{enumerate}
\end{definition}

According to the state of a clique, the issues in $T$ can be trisected into three parts.

\begin{definition}
For two thresholds $\beta_{\mathbb{N}}$ and $\alpha_{\mathbb{N}}$ satisfying $0 \leq \beta_{\mathbb{N}} \leq \alpha_{\mathbb{N}} \leq 1$, the alliance issues $T_{\mathbb{N}}^{=}(G)$, neutral issues $T_{\mathbb{N}}^{\approx}(G)$, and conflict issues $T_{\mathbb{N}}^{\asymp}(G)$ of a clique $G \subseteq T$ based on the non-consistency measure $\mathbb{NM}$ are defined as:
\begin{align}
T_{\mathbb{N}}^{=}(G) &= \{ t \in T \mid \mathbb{NM}(G,t) \leq \beta_{\mathbb{N}} \}, \notag\\
T_{\mathbb{N}}^{\approx}(G) &= \{ t \in T \mid \beta_{\mathbb{N}} < \mathbb{NM}(G,t) < \alpha_{\mathbb{N}} \}, \\
T_{\mathbb{N}}^{\asymp}(G) &= \{ t \in T \mid \mathbb{NM}(G,t) \geq \alpha_{\mathbb{N}} \}.\notag
\end{align}
\end{definition}

For a clique $G$, the sets $T_{\mathbb{N}}^{=}(G)$, $T_{\mathbb{N}}^{\approx}(G)$, and $T_{\mathbb{N}}^{\asymp}(G)$ represent the sets of issues on which $G$ has a low, moderate, and high non-consistency degree, respectively. Apparently, they form a trisection of $T$, that is, we have: 
$T_{\mathbb{N}}^{=}(G) \cap T_{\mathbb{N}}^{\approx}(G) = \emptyset, 
T_{\mathbb{N}}^{=}(G) \cap T_{\mathbb{N}}^{\asymp}(G) = \emptyset, 
T_{\mathbb{N}}^{\approx}(G) \cap T_{\mathbb{N}}^{\asymp}(G) = \emptyset$, and 
$T_{\mathbb{N}}^{=}(G) \cup T_{\mathbb{N}}^{\approx}(G) \cup T_{\mathbb{N}}^{\asymp}(G) = T$.

\subsection{Selecting optimal feasible strategies with non-consistency measures}\label{4.2jie}\hspace{0.2in}
We present the method for identifying feasible strategies and optimal feasible strategies based on the non-consistency measures.

\begin{definition}\label{def_FS-N}
A strategy $J \in ST$ is called a feasible strategy for a clique $G \subseteq P$ regarding the non-consistency measure $\mathbb{NM}'$ if: 
\begin{align}
\frac{\#(G)}{\#(P)} \geq \gamma_P  \wedge \mathbb{NM}(G,J) \leq \tau,
\end{align}
where $\gamma_{P}\in[0,1]$ and $\tau \in [0, 0.5]$ are two given thresholds. The set of all feasible strategies for the clique $G$ obtained by the non-consistency measure $\mathbb{NM}'$ is denoted by $FS^{\mathbb{N}}(G)$.
\end{definition}

\begin{definition}\label{def_FS-LN}
An $L$-order strategy $J \in ST_{L}$ is called an $L$-order feasible strategy for a clique $G \subseteq P$ concerning the non-consistency measure $\mathbb{NM}'$ if: 
\begin{align}
\frac{\#(G)}{\#(P)} \geq \gamma_P  \wedge \mathbb{NM}(G,J) \leq \tau,
\end{align}
where $\gamma_{P}\in[0,1]$ and $\tau \in [0, 0.5]$ are two given thresholds. The set of all $L$-order feasible strategies for the clique $G$ obtained by the non-consistency measure $\mathbb{NM}'$ is denoted by $FS^{\mathbb{N}}_{L}(G)$.
\end{definition}

We select the optimal feasible strategies with non-consistency measures in a similar way as with consistency measures.

\begin{definition}\label{def_FS-Nopt} 
An optimal feasible strategy for a clique $G \subseteq P$ with respect to the non-consistency measure $\mathbb{NM}'$ is: 
\begin{align}\label{eq_FS-Nopt}
\widehat{J} \in \{ \argmin\limits_{J\in FS^{\mathbb{N}}(G) \wedge \frac{\#(J)}{\#(T)} \geq \gamma_{T}} \mathbb{NM}(G,J) \},
\end{align}
where $\gamma_T \in [0,1]$ is a given threshold. The set of all optimal feasible strategies for the clique $G$ obtained by non-consistency measure $\mathbb{NM}'$ is represented by $\widehat{FS}^{\mathbb{N}}(G)$.
\end{definition}

Equation \eqref{eq_FS-Nopt} expresses two conditions: (1) An optimal feasible strategy must be relatively large enough compared to $T$, and (2) it has the minimum non-consistency degree among all feasible strategies.
An $L$-order optimal feasible strategy can also be defined.

\begin{definition}\label{def_FS-LNopt} 
An $L$-order optimal feasible strategy for a clique $G \subseteq P$ regarding the non-consistency measure $\mathbb{NM}'$ is: 
\begin{align}\label{eq_FS-LNopt}
\widehat{J}_{L} \in \{ \argmin\limits_{J\in FS_{L}^{\mathbb{N}}(G) \wedge \frac{\#(J)}{\#(T)} \geq \gamma_{T}} \mathbb{NM}(G,J) \},
\end{align}
where $\gamma_T \in [0,1]$ is a given threshold. The set of all $L$-order optimal feasible strategies for the clique $G$ obtained by the non-consistency measure $\mathbb{NM}'$ is represented by $\widehat{FS}_{L}^{\mathbb{N}}(G)$.
\end{definition}

Algorithm \ref{algorithm3} is presented for selecting feasible strategies and $L$-order feasible strategies with non-consistency measures, and Algorithm \ref{algorithm4} for selecting optimal ones.
{\color{black}
In Algorithm \ref{algorithm3}, the time complexity of Lines 2-8 is $O((\#(G))^{2}\times\#(T))$, the time complexity of Lines 9-10 is $O(2^{\#(T)})$, and the time complexity of Lines 11-19 is $O(2^{\#(T)-1})$. Thus, the total time complexity of Algorithm \ref{algorithm3} is $O(\max\{(\#(G))^{2}\times\#(T),2^{\#(T)}\})$ and the total space complexity of Algorithm \ref{algorithm3} is $O(\#(T)\times2^{\#(T)})$.
In Algorithm \ref{algorithm4}, the time complexity of Lines 2-8 is $O(\#(FS^{\mathbb{N}}(G)))$ and the time complexity of Lines 9-15 is $O(\#(FS_{L}^{\mathbb{N}}(G)))$. Thus, the total time complexity of Algorithm \ref{algorithm4} is $O(\#(FS^{\mathbb{N}}(G)))$ and the total space complexity of Algorithm \ref{algorithm4} is $O(\#(T)\times\#(FS^{\mathbb{N}}(G)))$.
}

\begin{algorithm}[!ht]
	\setstretch{1.25}
	\caption{The algorithm of selecting the sets of feasible strategies $FS^{\mathbb{N}}(G)$ and $L$-order feasible strategies $FS_{L}^{\mathbb{N}}(G)$ for a clique $G$.}
	\label{algorithm3}
	\SetKwData{Left}{left}\SetKwData{This}{this}\SetKwData{Up}{up}
	\SetKwFunction{Union}{Union}\SetKwFunction{FindCompress}{FindCompress}
	\SetKwInOut{Input}{Input}\SetKwInOut{Output}{Output}
	
	\Input{A three-valued situation table $TS = (P, T, V, r)$; \\
		The target clique $G \subseteq P$ satisfying $\frac{\#(G)}{\#(P)}\geq\gamma_{P}$; \\
		The weight vector of agents $\Theta=(\theta(p_{1}),\theta(p_{2}),\dots,\theta(p_{m}))$; \\
		The weight vector of issues $W=(\omega(t_{1}),\omega(t_{2}),\dots,\omega(t_{n}))$; \\
		Parameters $\gamma_{P}\in [0,1]$, $\tau\in[0,0.5]$, and $L\in\{1,2,...,n\}$.}
	\Output{The sets of feasible strategies $FS^{\mathbb{N}}(G)$ and
		$L$-order feasible strategies $FS_{L}^{\mathbb{N}}(G)$ of $G$.}
	
	\Begin{		
        \For{each issue $t \in T$}{
	      \For{ agents $p,q \in G$}{
		    {\textbf{Compute}
			$\theta(p|pq) = \frac{\theta(p)}{\theta(p) + \theta(q)}$ and 
			$\theta(q|pq) = \frac{\theta(q)}{\theta(p) + \theta(q)}$ by Equation~\eqref{eq_theta-ab}}.
		
		    {\textbf{Compute} 
            $\mathbb{CA}_{t}(p,q) = |r(p,t)-r(q,t)| \times \min\{\theta(p|pq),\theta(q|pq)\}$ by Theorem~\ref{theorem4.1}}.	
	      }
	  {\textbf{Compute} 
        $\mathbb{NM}(G,t) = \frac{\sum\limits_{(p,q) \in G \times G}\mathbb{CA}_{t}(p,q)}{\left( \#(G) \right)^{2}}$ by Equation~\eqref{eq_NM-i}.}
        }

        \textbf{Initialize} $FS^{\mathbb{N}}(G) = \emptyset$ and $FS_{L}^{\mathbb{N}}(G) = \emptyset$.

        \textbf{Create} the set of strategies $ST = \{J\in T \mid J \neq \emptyset\}$.
		
		\For{each strategy $J\in ST$}{
			
			{\textbf{Compute} 
            $\mathbb{NM}(G,J) = \sum\limits_{t \in J}\big(\omega(t|J)\times \mathbb{NM}(G,t)\big)$ by Equation \eqref{eq_NM-J}}. 
			
			\If{$\frac{\#(G)}{\#(P)}\geq\gamma_{T}$ and $\mathbb{NM}(G,J)\leq\tau$}{
				$FS^{\mathbb{N}}(G) = FS^{\mathbb{N}}(G) \cup \{J\}$ by Definition~\ref{def_FS-N}.
				
				\If{$\#(J)=L$}{
					$FS_{L}^{\mathbb{N}}(G) = FS_{L}^{\mathbb{N}}(G) \cup \{J\}$ by Definition~\ref{def_FS-LN}.
				}
			}
		}
	}
	\textbf{Return:} $FS^{\mathbb{N}}(G)$ and $FS_{L}^{\mathbb{N}}(G)$.
\end{algorithm}

\begin{algorithm}[!ht]
	\setstretch{1.25}
	\caption{The algorithm of selecting the sets of optimal feasible strategies $\widehat{FS}^{\mathbb{N}}(G)$ and $L$-order optimal feasible strategies $\widehat{FS}_{L}^{\mathbb{N}}(G)$ for a clique $G$.}
	\label{algorithm4}
	\SetKwData{Left}{left}\SetKwData{This}{this}\SetKwData{Up}{up}
	\SetKwFunction{Union}{Union}\SetKwFunction{FindCompress}{FindCompress}
	\SetKwInOut{Input}{Input}\SetKwInOut{Output}{Output}
	
	\Input{The set of feasible strategies $FS^{\mathbb{N}}(G)$ of $G$; \\
		The non-consistency degree of each strategy in $FS^{\mathbb{N}}(G)$; \\
		The set of $L$-order feasible strategies $FS_{L}^{\mathbb{N}}(G)$ of $G$; \\
		The non-consistency degree of each $L$-order strategy in $FS_{L}^{\mathbb{N}}(G)$; \\
        Parameter $\gamma_T \in [0,1]$.}
	\Output{The set of optimal feasible strategies $\widehat{FS}^{\mathbb{N}}(G)$ of $G$;\\
		The set of $L$-order optimal feasible strategies $\widehat{FS}_{L}^{\mathbb{N}}(G)$ of $G$.}
	
	\Begin{
		\textbf{Compute} the minimum non-consistency degree 
		$\mathbb{NM}^{min} = \min\limits_{J\in FS^{\mathbb{N}}(G)}\mathbb{NM}(G,J)$. \\
		
		\textbf{Initialize} $\widehat{FS}^{\mathbb{N}}(G) = \emptyset$.
		
		\For{each feasible strategy $J \in FS^{\mathbb{N}}(G)$}{
			\If{$\frac{\#(J)}{\#(T)} \geq \gamma_T$ and $\mathbb{NM}(G,J) = \mathbb{NM}^{min}$}{
				$\widehat{FS}^{\mathbb{N}}(G) = \widehat{FS}^{\mathbb{N}}(G) \cup \{J\}$ by Definition \ref{def_FS-Nopt}.
			}
		}
		
		\textbf{Compute} the minimum non-consistency degree 
		$\mathbb{NM}^{min}_{L} = \min\limits_{J_{L}\in FS_{L}^{\mathbb{N}}(G)}\mathbb{NM}(G,J)$. \\
		
		\textbf{Initialize} $\widehat{FS}_{L}^{\mathbb{N}}(G) = \emptyset$.
		
		\For{each $L$-order feasible strategy $J_{L}\in FS_{L}^{\mathbb{N}}(G)$}{
			\If{$\frac{\#(J_{L})}{\#(T)} \geq \gamma_T$ and $\mathbb{NM}(G,J_{L}) = \mathbb{NM}^{min}_{L}$}{
				$\widehat{FS}_{L}^{\mathbb{N}}(G) = \widehat{FS}_{L}^{\mathbb{N}}(G) \cup \{J_{L}\}$ by Definition \ref{def_FS-LNopt}.
			}
		}	
	}
\textbf{Return:} $\widehat{FS}^{\mathbb{N}}(G)$ and $\widehat{FS}_{L}^{\mathbb{N}}(G)$.
\end{algorithm}

\begin{example}\textbf{(Continuing with Example~\ref{ex4.2})}
We illustrate the above discussion with non-consistency measures considering the clique $G = \{p_{1},p_{3},p_{4},p_{6}\}$. Table \ref{tab_ST-N} gives all strategies and their non-consistency degrees.
Firstly, with $\tau = 0.27$, by Definition \ref{def_FS-N}, we obtain the set of feasible strategies for the clique $G$ as: 
\begin{align*}
FS^{\mathbb{N}}(G) = \{ \{t_1\}, \{t_4\}, \{t_1, t_4\}, \{t_1, t_3, t_4\} \}.
\end{align*}
By Definition \ref{def_FS-LN}, we obtain the set of $3$-order feasible strategies for $G$ as:
\begin{align*}
FS_{3}^{\mathbb{N}}(G) = \{ \{t_1, t_3, t_4\} \}.  
\end{align*}
Finally, with $\gamma_T = 0.5$, by Definitions \ref{def_FS-Nopt} and \ref{def_FS-LNopt}, we determine the sets of optimal feasible strategy and $3$-order optimal feasible strategy for $G$ as:  
\begin{align*}
\widehat{FS}^{\mathbb{N}}(G) = \widehat{FS}_{3}^{\mathbb{N}}(G) = \{ \{t_1, t_3, t_4\} \}.
\end{align*}
In Table~\ref{tab_ST-N}, we use the same highlights as in Table \ref{tab_ST-C}.
Combining Tables \ref{tab_ST-C} and \ref{tab_ST-N}, the optimal feasible strategy and $L$-order optimal ones derived from the consistency measure is $\{t_1,t_4,t_5\}$, whereas the optimal feasible strategy and $L$-order optimal ones obtained through the non-consistency measure is $\{t_1,t_3,t_4\}$. Although the two results are not exactly the same, they are highly coincident.

\begin{table}[!ht]
	\renewcommand{\arraystretch}{1.3}
	\begin{center}
		\caption{Strategies and their non-consistency degrees with respect to $G$.}\label{tab_ST-N}
		\setlength{\tabcolsep}{5.3mm}
		\begin{tabular}{lc|lc|lc}
			\hline
			$J$  & $\mathbb{NM}(G,J)$ & $J$ & $\mathbb{NM}(G,J)$ & $J$ & $\mathbb{NM}(G,J)$ \\
			\hline
			\cellcolor{green!20}$\{t_{1}\}$ & \cellcolor{green!20}$0.2655$ 
			& $\{t_{2},t_{5}\}$ & $0.3374$   
			& $\{t_{2},t_{3},t_{5}\}$ & $0.3096$ \\
			
			$\{t_{2}\}$ & $0.3557$ 
			& $\{t_{3},t_{4}\}$ & $0.2723$ 
			& $\{t_{2},t_{4},t_{5}\}$  & $0.3134$ \\
			
			$\{t_{3}\}$ & $0.2763$ 
			& $\{t_{3},t_{5}\}$ & $0.2994$  
			& $\{t_{3},t_{4},t_{5}\}$ & $0.2909$ \\
			
			\cellcolor{green!20}$\{t_{4}\}$ & \cellcolor{green!20}$0.2655$ 
			& $\{t_{4},t_{5}\}$ & $0.3014$ 
			& $\{t_{1},t_{2},t_{3},t_{4}\}$ & $0.2802$ \\
			
			$\{t_{5}\}$ & $0.3283$ 
			& $\{t_{1},t_{2},t_{3}\}$ & $0.2835$
			& $\{t_{1},t_{2},t_{3},t_{5}\}$ & $0.2941$ \\
			
			$\{t_{1},t_{2}\}$ & $0.2881$ 
			& $\{t_{1},t_{2},t_{4}\}$ & $0.2819$ 
			& $\{t_{1},t_{2},t_{4},t_{5}\}$ & $0.2943$ \\
			
			$\{t_{1},t_{3}\}$ & $0.2704$ 
			& $\{t_{1},t_{2},t_{5}\}$ & $0.3015$
			& $\{t_{1},t_{3},t_{4},t_{5}\}$ & $0.2825$ \\
			
			\cellcolor{green!20}$\{t_{1},t_{4}\}$ & \cellcolor{green!20}$0.2655$ 
			& \cellcolor{green!20}\color{blue}\underline{$\{t_{1},t_{3},t_{4}\}$} 
			& \cellcolor{green!20}\color{blue}\underline{$0.2694$} 
			& $\{i_{2},i_{3},i_{4},i_{5}\}$ & $0.3002$\\
			
			$\{t_{1},t_{5}\}$ & $0.2906$ 
			& $\{t_{1},t_{3},t_{5}\}$ & $0.2858$
			& $\{t_{1},t_{2},t_{3},t_{4},t_{5}\}$ & $0.2898$\\
			
			$\{t_{2},t_{3}\}$ & $0.2990$ 
			& $\{t_{1},t_{4},t_{5}\}$ & $0.2848$ \\
			
			$\{t_{2},t_{4}\}$ & $0.3016$ 
			& $\{t_{2},t_{3},t_{4}\}$ & $0.2889$ \\
			\hline
		\end{tabular}
	\end{center}
\end{table}
\end{example}

\section{Case studies, sensitivity, and comparative analysis}\label{5jie}\hspace{0.2in}
In this section, we apply the two proposed models to real-world scenarios. Additionally, we conduct a sensitivity analysis of all parameters and provide a comparative analysis of these two models against eight existing relevant conflict analysis models.

\subsection{Case studies}\label{5.1jie}\hspace{0.2in}
Our proposed models are applied to the NBA labor negotiations and the development plans for Gansu Province to validate their practicality.

\subsubsection{Case studies in NBA labor negotiations}\label{5.11jie}\hspace{0.2in}
We employ our proposed models to evaluate the conflict situations in the NBA labor negotiations, {\color{black}which are used as case studies by Xu \cite{Xu2022}.} Moreover, we also propose suitable feasible strategies for conflict resolution.

Table~\ref{tab_NBA}~\cite{Xu2022} illustrates the conflict dynamics within the NBA labor negotiations, where the set of agents $P = \{ p_1, p_2, \dots, p_{10} \}$ includes the president, representatives of the team owners, lawyers, financial experts, the chairman and executive director of the players' union, and representatives from the negotiating committee, in which the negotiating committee consists of players from various income brackets and seniority levels. The set of issues $T = \{ t_1, t_2, \dots, t_9 \}$ encompasses critical aspects, such as total player wages, minimum work standards, club owner commissions, the ratio of player income to league revenue, the total number of players hired by each team, age restrictions for hired players, working conditions, insurance provided to players, and the proportion of foreign players on each team. The ratings ``$+1$'' and ``$-1$'' are again omitted as ``$+$'' and ``$-$'', respectively.

\begin{table}[!ht]
\renewcommand{\arraystretch}{1.3}
\begin{center}
\caption{The conflict situation in the NBA labor negotiation~\cite{Xu2022}.}
\label{tab_NBA}
\setlength{\tabcolsep}{6.4mm}
\begin{tabular}{cccccccccc}
\hline
& $t_1$ & $t_2$ & $t_3$ & $t_4$ & $t_5$ & $t_6$ & $t_7$ & $t_8$ & $t_9$ \\
\hline
$p_1$    & $+$ & $+$ & $+$ & $-$ & $0$ & $-$ & $-$ & $-$ & $+$ \\
$p_2$    & $+$ & $+$ & $-$ & $+$ & $-$ & $0$ & $-$ & $+$ & $-$ \\
$p_3$    & $+$ & $0$ & $+$ & $+$ & $+$ & $-$ & $-$ & $-$ & $0$ \\
$p_4$    & $+$ & $+$ & $+$ & $-$ & $-$ & $+$ & $+$ & $0$ & $-$ \\
$p_5$    & $-$ & $+$ & $+$ & $+$ & $+$ & $-$ & $0$ & $-$ & $-$ \\
$p_6$    & $+$ & $+$ & $-$ & $0$ & $-$ & $+$ & $-$ & $+$ & $0$ \\
$p_7$    & $+$ & $+$ & $+$ & $+$ & $+$ & $-$ & $+$ & $0$ & $-$ \\
$p_8$    & $+$ & $-$ & $+$ & $-$ & $0$ & $+$ & $-$ & $-$ & $-$ \\
$p_9$    & $+$ & $+$ & $0$ & $+$ & $+$ & $+$ & $-$ & $-$ & $-$ \\
$p_{10}$ & $-$ & $+$ & $+$ & $-$ & $+$ & $0$ & $0$ & $-$ & $-$ \\
\hline
\end{tabular}
\end{center}
\end{table}

In the NBA labor negotiations, the ten agents possess varying levels of influence, while the nine issues differ in terms of their importance. {\color{black}Similar to the examples in Sections \ref{3jie} and \ref{4jie}, based on expert judgment and domain experience, the weight vectors of agents and issues are given by $\Theta = (0.25, 0.18, 0.07, 0.06, 0.05, 0.1, 0.09, 0.12, \\ 0.03, 0.05)$ and $W = (0.3, 0.03, 0.02, 0.2, 0.15, 0.1, 0.1, 0.05, 0.05)$, respectively.}
With $\gamma_P = 0.5$, we take a clique $G = \{p_1, p_2, p_3, p_6, p_9\}$. Then we derive the following six types of strategies, given in Table \ref{tab_FS}.

\begin{enumerate}[label = (\arabic*)]
\setlength{\itemsep}{0pt}
\item By Definition~\ref{def_ST}, we identify 511 strategies for the clique $G$. Due to the large number of these strategies, they are not listed in Table~\ref{tab_FS}.

\item For $L = 5$, we apply Definition~\ref{def_ST-L} to obtain 126 5-order strategies for the clique $G$, which are again not listed in Table~\ref{tab_FS}.

\item For $\mu = 0.3$ and $\nu = -0.3$, we calculate the vector of overall ratings of the clique $G$ toward all issues using Equation~\eqref{eq_At}, resulting in $\bm{\mathrm{R}}_{G} = (+1, +1, 0, 0, 0, -1, -1, 0, 0)$. 
Then, by setting $\lambda = 0.94$, we apply Algorithm~\ref{algorithm1} to obtain 19 consistency-based feasible strategies for the clique $G$, which form the set of feasible strategies $FS^{\mathbb{C}}(G)$ in Table~\ref{tab_FS}.

\item By Definition \ref{def_FS-LC}, we obtain the set of $5$-order feasible strategies of $G$ based on the consistency measure, namely $FS_{5}^{\mathbb{C}}(G)$ in Table \ref{tab_FS}.

\item With $\tau = 0.06$, we use Algorithm~\ref{algorithm3} to obtain 33 non-consistency-based feasible strategies for the clique $G$, forming the set $FS^{\mathbb{N}}(G)$, as presented in Table~\ref{tab_FS}. 

\item By Definition \ref{def_FS-LN}, we obtain four $5$-order non-consistency-based feasible strategies of $G$, forming the set $FS_{5}^{\mathbb{N}}(G)$ in Table \ref{tab_FS}.
\end{enumerate}

\begin{table}[!ht]
\renewcommand{\arraystretch}{1.3}
\begin{center}
\caption{The six types of strategies for the clique $G$ in NBA labor negotiations.}
\label{tab_FS}
\setlength{\tabcolsep}{3.55mm}
\begin{tabular}{lll}
\hline
Type & Strategies & Number  \\

\hline
~$ST$     &  $\{J\subseteq T \mid J\neq\emptyset\}$ & 511 \\
\rule{0pt}{16pt}
$ST_{5}$ &  $\{J\subseteq ST \mid \#(J)=5\}$ & 126 \\
\rule{0pt}{28pt}
$FS^{\mathbb{C}}(G)$ & \makecell[l]{
\{$\{t_{1}\}$,$\{t_{2}\}$,$\{t_{7}\}$,$\{t_{1},t_{2}\}$,$\{t_{1},t_{3}\}$,$\{t_{1},t_{7}\}$,$\{t_{1},t_{9}\}$,$\{t_{2},t_{7}\}$,$\{t_{1},t_{2},t_{3}\}$,$\{t_{1},t_{2},t_{7}\}$,\\
$\{t_{1},t_{2},t_{9}\}$,~\,$\{t_{1},t_{3},t_{7}\}$,~~$\{t_{1},t_{7},t_{8}\}$,~~$\{t_{1},t_{7},t_{9}\}$,~~$\{t_{1},t_{2},t_{3},t_{7}\}$,~~$\{t_{1},t_{2},t_{7},t_{8}\}$,\\
$\{t_{1},t_{2},t_{7},t_{9}\}$,$\{t_{1},t_{3},t_{7},t_{9}\}$,\underline{$\{t_{1},t_{2},t_{3},t_{7},t_{9}\}$}\}} & 19 \\
\rule{0pt}{16pt}
$FS_{5}^{\mathbb{C}}(G)$ & \makecell[l]{\{\underline{$\{t_{1},t_{2},t_{3},t_{7},t_{9}\}$}\}} & 1 \\
\rule{0pt}{48pt}
$FS^{\mathbb{N}}(G)$ & \makecell[l]{
\{$\{t_{1}\}$,~~$\{t_{7}\}$,~~$\{t_{1},t_{2}\}$,~~$\{t_{1},t_{3}\}$,~~$\{t_{1},t_{7}\}$,~~$\{t_{1},t_{8}\}$,~~\,$\{t_{1},t_{9}\}$,~~\,$\{t_{2},t_{7}\}$,~~\,$\{t_{3},t_{7}\}$,\\
$\{t_{1},t_{2},t_{3}\}$,$\{t_{1},t_{2},t_{7}\}$,$\{t_{1},t_{2},t_{8}\}$,$\{t_{1},t_{2},t_{9}\}$,$\{t_{1},t_{3},t_{7}\}$,\,$\{t_{1},t_{3},t_{8}\}$,\,$\{t_{1},t_{3},t_{9}\}$,\\
$\{t_{1},t_{6},t_{7}\}$,~\,$\{t_{1},t_{7},t_{8}\}$,~\,$\{t_{1},t_{7},t_{9}\}$,~~$\{t_{2},t_{3},t_{7}\}$,~~$\{t_{1},t_{2},t_{3},t_{7}\}$,~~$\{t_{1},t_{2},t_{3},t_{8}\}$,\\
$\{t_{1},t_{2},t_{3},t_{9}\}$,~~~$\{t_{1},t_{2},t_{6},t_{7}\}$,~~~$\{t_{1},t_{2},t_{7},t_{8}\}$,~~~$\{t_{1},t_{2},t_{7},t_{9}\}$,~~~\,$\{t_{1},t_{3},t_{7},t_{8}\}$,\\
$\{t_{1},t_{3},t_{7},t_{9}\}$,~$\{t_{1},t_{7},t_{8},t_{9}\}$,~$\{t_{1},t_{2},t_{3},t_{7},t_{8}\}$,~\underline{$\{t_{1},t_{2},t_{3},t_{7},t_{9}\}$},~$\{t_{1},t_{2},t_{7},t_{8},$\\
$t_{9}\}$,$\{t_{1},t_{3},t_{7},t_{8},t_{9}\}$\}} 
& 33\\
\rule{0pt}{16pt}
$FS_{5}^{\mathbb{N}}(G)$ & \makecell[l]{\{$\{t_{1},t_{2},t_{3},t_{7},t_{8}\}$, ~\underline{$\{t_{1},t_{2},t_{3},t_{7},t_{9}\}$},$\{t_{1},t_{2},t_{7},t_{8},t_{9}\}$,$\{t_{1},t_{3},t_{7},t_{8},t_{9}\}$\}} & 4 \\
\hline
\end{tabular}
\end{center}
\end{table}

With $\gamma_T = 0.5$, as $\#(J) \geq \gamma_T \times \#(T) = 0.5 \times 9 = 4.5$, the optimal feasible strategy for the clique $G$ must involve at least five issues. By Algorithms \ref{algorithm2} and \ref{algorithm4}, we obtain the set of optimal feasible strategies of $G$, specifically 
$\widehat{FS}^{\mathbb{C}}(G) = \widehat{FS}_{5}^{\mathbb{C}}(G) = \widehat{FS}^{\mathbb{N}}(G) = \widehat{FS}_{5}^{\mathbb{N}}(G) = \{ \{t_1, t_2, t_3, t_7, t_9\} \}$, 
which are underlined in Table~\ref{tab_FS}.
As demonstrated in Table~\ref{tab_FS}, it is noteworthy that the optimal feasible strategies derived from the consistency and non-consistency perspectives are the same. 
In summary, the proposed models facilitate the systematic selection of feasible strategies from perspectives of consistency and non-consistency, validating their practical applicability in the real-world NBA labor negotiation.

{\color{black}
\subsubsection{Case studies in development plans for Gansu Province}\label{5.12jie}\hspace{0.2in}
Our proposed models are further applied to analyze conflict situations in making development plans for Gansu Province, which is a case previously examined by Sun \cite{Sun2020}. Additionally, we develop actionable conflict resolution strategies tailored to this case.

The conflict situation in making development plans of Gansu Province is characterized in Table \ref{tab_GS} \cite{Sun2020}, where $P = \{p_{1}, p_{2}, \dots, p_{14}\}$ denotes the set of fourteen cities in Gansu Province, namely Lanzhou, Jinchang, Baiyin, Tianshui, Jiayuguan, Wuwei, Zhangye, Pingliang, Jiuquan, Qingyang, Dingxi, Longnan, Linxia, and Gannan;
$T = \{t_{1}, t_{2}, ..., t_{11}\}$ represents the set of eleven issues, including the construction of roads, factories, entertainment, educational institutions, the total population of residence, ecology environment, the number of senior intellectuals, the traffic capacity, mineral resources, sustainable development capacity, and water resources carrying capacity. The ratings ``$+1$'' and ``$-1$'' are hereafter abbreviated as ``$+$'' and ``$-$'', respectively.

\begin{table}[!ht]
\renewcommand{\arraystretch}{1.3}
\begin{center}
\caption{The conflict situation in making development plans of Gansu Province~\cite{Sun2020}.}
\label{tab_GS}
\setlength{\tabcolsep}{5.0mm}
\begin{tabular}{cccccccccccc}
\hline
& $t_1$ & $t_2$ & $t_3$ & $t_4$ & $t_5$ & $t_6$ & $t_7$ & $t_8$ & $t_9$ & $t_{10}$ & $t_{11}$ \\
\hline
$p_1$    & $+$ & $-$ & $0$ & $-$ & $+$ & $-$ & $0$ & $-$ & $+$ & $-$ & $+$ \\
$p_2$    & $0$ & $+$ & $-$ & $0$ & $0$ & $+$ & $-$ & $0$ & $0$ & $+$ & $-$ \\
$p_3$    & $-$ & $0$ & $-$ & $-$ & $-$ & $+$ & $+$ & $-$ & $-$ & $0$ & $0$ \\
$p_4$    & $0$ & $0$ & $-$ & $+$ & $+$ & $-$ & $-$ & $+$ & $0$ & $-$ & $-$ \\
$p_5$    & $-$ & $+$ & $-$ & $0$ & $-$ & $+$ & $0$ & $0$ & $-$ & $+$ & $+$ \\
$p_6$    & $0$ & $+$ & $0$ & $-$ & $-$ & $-$ & $-$ & $-$ & $0$ & $+$ & $-$ \\
$p_7$    & $+$ & $+$ & $0$ & $+$ & $0$ & $+$ & $0$ & $+$ & $+$ & $+$ & $0$ \\
$p_8$    & $-$ & $0$ & $-$ & $+$ & $-$ & $0$ & $+$ & $+$ & $-$ & $0$ & $+$ \\
$p_9$    & $+$ & $+$ & $0$ & $-$ & $+$ & $+$ & $-$ & $-$ & $+$ & $+$ & $-$ \\
$p_{10}$ & $-$ & $-$ & $-$ & $0$ & $+$ & $-$ & $+$ & $0$ & $-$ & $-$ & $+$ \\
$p_{11}$ & $-$ & $0$ & $-$ & $-$ & $-$ & $-$ & $-$ & $-$ & $-$ & $0$ & $-$ \\
$p_{12}$ & $0$ & $+$ & $0$ & $-$ & $+$ & $+$ & $+$ & $-$ & $0$ & $+$ & $0$ \\
$p_{13}$ & $-$ & $0$ & $-$ & $+$ & $0$ & $0$ & $0$ & $+$ & $-$ & $-$ & $+$ \\
$p_{14}$ & $-$ & $-$ & $-$ & $0$ & $-$ & $-$ & $-$ & $0$ & $-$ & $0$ & $-$ \\
\hline
\end{tabular}
\end{center}
\end{table}

In making development plans for Gansu Province, we assume that the weight vector for the agents is $\Theta = (0.18,0.05,0.06,0.08,0.05,0.07,0.05,0.05,0.1,0.15,0.05,0.05,0.03,0.03)$, and the weight vector for the issues is $W = (0.08,0.07,0.03,0.12,0.05,0.13,0.1,0.1,0.05,0.12,0.15)$.
The parameters are set as follows: $\gamma_P = \gamma_T = 0.5$, $\mu = 0.3$, $\nu = -0.3$, $\lambda = 0.72$, $\tau = 0.275$, and $L = 6$. For a clique $G = \{p_1, p_3, p_4, p_5, p_6, p_9, p_{10}, p_{11}\}$, six distinct strategy types are derived, as detailed in Table~\ref{tab_FS-GS}.
By applying Algorithms \ref{algorithm2} and \ref{algorithm4}, we identify the set of optimal feasible strategies for clique $G$. Specifically,  
the set of consistency-based optimal feasible strategies is $\widehat{FS}^{\mathbb{C}}(G) = \widehat{FS}_{6}^{\mathbb{C}}(G) = \{ \{t_3, t_4, t_5, t_6, t_8, t_9\} \}$ and the set of non-consistency-based optimal feasible strategies is $\widehat{FS}^{\mathbb{N}}(G) = \widehat{FS}_{6}^{\mathbb{N}}(G) = \{ \{t_2, t_3, t_4, t_5, t_8, t_9\} \}$. 
These results are underlined in Table \ref{tab_FS-GS} for clarity. 
Notably, the optimal strategies obtained from both consistency and inconsistency perspectives exhibit remarkable convergence. Although these approaches originate from contrasting theoretical foundations, they converge on nearly identical solutions. This phenomenon has also been observed in case studies such as the Middle East conflict and NBA labor negotiations, further attesting to the practical applicability of our model.

\begin{table}[!ht]
\renewcommand{\arraystretch}{1.3}
\begin{center}
\caption{The six types of strategies for the clique $G$ in the development plans of Gansu Province.}
\label{tab_FS-GS}
\setlength{\tabcolsep}{3.05mm}
\begin{tabular}{lll}
\hline
Type & Strategies & Number  \\

\hline
~$ST$     &  $\{J\subseteq T \mid J\neq\emptyset\}$ & 2047 \\

\rule{0pt}{16pt}
$ST_{6}$ &  $\{J\subseteq ST \mid \#(J)=6\}$ & 462 \\

\rule{0pt}{64pt}
$FS^{\mathbb{C}}(G)$ & \makecell[l]{
\{$\{t_{3}\}$,$\{t_{4}\}$,$\{t_{8}\}$,$\{t_{3},t_{4}\}$,$\{t_{3},t_{6}\}$,$\{t_{3},t_{8}\}$,$\{t_{4},t_{5}\}$,$\{t_{4},t_{6}\}$,$\{t_{4},t_{8}\}$,$\{t_{5},t_{8}\}$,$\{t_{6},t_{8}\}$,\\
$\{t_{2},t_{4},t_{8}\}$,$\{t_{3},t_{4},t_{5}\}$,~$\{t_{3},t_{4},t_{6}\}$,~$\{t_{3},t_{4},t_{8}\}$,~$\{t_{3},t_{5},t_{8}\}$,~$\{t_{3},t_{6},t_{8}\}$,~$\{t_{4},t_{5},t_{6}\}$,\\
$\{t_{4},t_{5},t_{8}\}$,~~~\,$\{t_{4},t_{6},t_{8}\}$,~~~\,$\{t_{4},t_{7},t_{8}\}$,~~~\,$\{t_{4},t_{8},t_{9}\}$,~~~\,$\{t_{5},t_{6},t_{8}\}$,~~~\,$\{t_{2},t_{3},t_{4},t_{8}\}$,\\
$\{t_{2},t_{4},t_{5},t_{8}\}$,~~~~$\{t_{2},t_{4},t_{6},t_{8}\}$,~~~~$\{t_{3},t_{4},t_{5},t_{6}\}$,~~~~$\{t_{3},t_{4},t_{5},t_{8}\}$,~~~~$\{t_{3},t_{4},t_{6},t_{8}\}$,\\
$\{t_{3},t_{4},t_{7},t_{8}\}$,~~~~$\{t_{3},t_{4},t_{8},t_{9}\}$,~~~~$\{t_{3},t_{5},t_{6},t_{8}\}$,~~~~$\{t_{4},t_{5},t_{6},t_{8}\}$,~~~~$\{t_{4},t_{5},t_{7},t_{8}\}$,\\
$\{t_{4},t_{5},t_{8},t_{9}\}$,~$\{t_{4},t_{6},t_{7},t_{8}\}$,~\,$\{t_{4},t_{6},t_{8},t_{9}\}$,~\,$\{t_{2},t_{3},t_{4},t_{5},t_{8}\}$,~\,$\{t_{2},t_{3},t_{4},t_{6},t_{8}\}$,\\
$\{t_{3},t_{4},t_{5},t_{6},t_{8}\}$,\,$\{t_{3},t_{4},t_{5},t_{7},t_{8}\}$,\,$\{t_{3},t_{4},t_{5},t_{8},t_{9}\}$,\,$\{t_{3},t_{4},t_{6},t_{7},t_{8}\}$,\,$\{t_{3},t_{4},t_{6},$\\
$t_{8},t_{9}\}$,$\{t_{4},t_{5},t_{6},t_{8},t_{9}\}$,$\{t_{2},t_{3},t_{4},t_{5},t_{6},t_{8}\}$,$\{t_{3},t_{4},t_{5},t_{6},t_{7},t_{8}\}$,\underline{$\{t_{3},t_{4},t_{5},t_{6},t_{8},$}\\
\underline{$t_{9}\}$}\}
} 
& 48 \\

\rule{0pt}{18pt}
$FS_{6}^{\mathbb{C}}(G)$ & \makecell[l]{\{$\{t_{2},t_{3},t_{4},t_{5},t_{6},t_{8}\}$,$\{t_{3},t_{4},t_{5},t_{6},t_{7},t_{8}\}$,\underline{$\{t_{3},t_{4},t_{5},t_{6},t_{8},t_{9}\}$}\}} & 3 \\

\rule{0pt}{64pt}
$FS^{\mathbb{N}}(G)$ & \makecell[l]{
\{$\{t_{3}\}$,$\{t_{4}\}$,$\{t_{8}\}$,$\{t_{1},t_{3}\}$,$\{t_{2},t_{3}\}$,$\{t_{2},t_{4}\}$,$\{t_{2},t_{8}\}$,$\{t_{3},t_{4}\}$,$\{t_{3},t_{5}\}$,$\{t_{3},t_{8}\}$,$\{t_{3},t_{9}\}$,\\
$\{t_{4},t_{8}\}$,~\,$\{t_{1},t_{2},t_{3}\}$,~\,$\{t_{1},t_{3},t_{4}\}$,~\,$\{t_{1},t_{3},t_{8}\}$,~\,$\{t_{1},t_{4},t_{8}\}$,~\,$\{t_{2},t_{3},t_{4}\}$,~\,$\{t_{2},t_{3},t_{8}\}$,\\
$\{t_{2},t_{3},t_{9}\}$,\,$\{t_{2},t_{4},t_{8}\}$,\,$\{t_{3},t_{4},t_{5}\}$,\,$\{t_{3},t_{4},t_{8}\}$,~$\{t_{3},t_{4},t_{9}\}$,~$\{t_{3},t_{5},t_{8}\}$,~$\{t_{3},t_{8},t_{9}\}$,\\
$\{t_{4},t_{5},t_{8}\}$,\,$\{t_{4},t_{8},t_{9}\}$,\,$\{t_{1},t_{2},t_{3},t_{4}\}$,\,$\{t_{1},t_{2},t_{3},t_{8}\}$,\,$\{t_{1},t_{3},t_{4},t_{8}\}$,\,$\{t_{1},t_{3},t_{4},t_{9}\}$,\\
$\{t_{1},t_{3},t_{8},t_{9}\}$,~~~~$\{t_{2},t_{3},t_{4},t_{5}\}$,~~~~$\{t_{2},t_{3},t_{4},t_{8}\}$,~~~~$\{t_{2},t_{3},t_{4},t_{9}\}$,~~~~$\{t_{2},t_{3},t_{5},t_{8}\}$,\\
$\{t_{2},t_{3},t_{8},t_{9}\}$,~~~~$\{t_{2},t_{4},t_{8},t_{9}\}$,~~~~$\{t_{3},t_{4},t_{5},t_{8}\}$,~~~~$\{t_{3},t_{4},t_{5},t_{9}\}$,~~~~$\{t_{3},t_{4},t_{7},t_{8}\}$,\\
$\{t_{3},t_{4},t_{8},t_{9}\}$,$\{t_{3},t_{5},t_{8},t_{9}\}$,$\{t_{1},t_{2},t_{3},t_{4},t_{8}\}$,\,$\{t_{1},t_{3},t_{4},t_{5},t_{8}\}$,\,$\{t_{1},t_{3},t_{4},t_{8},t_{9}\}$,\\
$\{t_{2},t_{3},t_{4},t_{5},t_{8}\}$,\,$\{t_{2},t_{3},t_{4},t_{7},t_{8}\}$,\,$\{t_{2},t_{3},t_{4},t_{8},t_{9}\}$,\,$\{t_{3},t_{4},t_{5},t_{8},t_{9}\}$,\,$\{t_{1},t_{2},t_{3},$\\
$t_{4},t_{5},t_{8}\}$,$\{t_{1},t_{2},t_{3},t_{4},t_{8},t_{9}\}$,\underline{$\{t_{2},t_{3},t_{4},t_{5},t_{8},t_{9}\}$}\}
}
& 53\\

\rule{0pt}{16pt}
$FS_{6}^{\mathbb{N}}(G)$ & \makecell[l]{\{$\{t_{1},t_{2},t_{3},t_{4},t_{5},t_{8}\}$,$\{t_{1},t_{2},t_{3},t_{4},t_{8},t_{9}\}$,\underline{$\{t_{2},t_{3},t_{4},t_{5},t_{8},t_{9}\}$}\}} & 3 \\
\hline
\end{tabular}
\end{center}
\end{table}

}

{\color{black}
\subsection{Sensitivity analysis}\label{5.2jie}\hspace{0.2in}
In this section, we perform a comprehensive sensitivity analysis based on the case of NBA labor negotiations to assess the impact of each parameter on the selection of feasible strategies.

\subsubsection{\texorpdfstring{Sensitivity analysis on parameters $\mu$ and $\nu$}
{Sensitivity analysis on parameters mu and nu}}\label{5.21jie}\hspace{0.2in}
Since the parameters 
$\mu$ and $\nu$ are unique to our consistency-based model, we systematically investigate their effects on the overall rating, consistency degree, and the set of consistency-based feasible strategies.

For the clique \(G = \{p_1, p_2, p_3, p_6, p_9\}\), with parameters set to \(\lambda = 0.94\) and \(L = 5\), we analyze the effects of varying \(\mu\) from 0 to 1 in increments of 0.2, and \(\nu\) from 0 to -1 in decrements of 0.2. The metrics examined under these parameter variations include the overall ratings of \(G\) on a fixed issue, the consistency degrees of \(G\) on that issue, and the number of distinct types of consistency-based feasible strategies. Due to the large number of issues, we focus on issue \(t_5\) as a representative example to illustrate in detail the overall rating and consistency degree of clique \(G\) across different parameter combinations. The full results of this sensitivity analysis for parameters \(\mu\) and \(\nu\) are presented in Tables \ref{tab_R-munu}–\ref{tab_opt-munu}.

\begin{table}[!ht]
\centering
\begin{minipage}[!ht]{0.45\textwidth}
\renewcommand{\arraystretch}{1.3}
\centering
\caption{The overall ratings for the clique $G$ regarding the issue $t_5$ under different values of $\mu$ and $\nu$.}
\label{tab_R-munu}
\setlength{\tabcolsep}{1.95mm}
\begin{tabular}{ccccccc}
\hline
\diagbox{$\mu$}{$\nu$} & $0$ & $-0.2$ & $-0.4$ & $-0.6$ & $-0.8$ & $-1$ \\ 
\hline
$0$   & $-$ & $-$ & $0$ & $0$ & $0$ & $0$ \\ 
$0.2$ & $-$ & $-$ & $0$ & $0$ & $0$ & $0$ \\  
$0.4$ & $-$ & $-$ & $0$ & $0$ & $0$ & $0$ \\ 
$0.6$ & $-$ & $-$ & $0$ & $0$ & $0$ & $0$ \\ 
$0.8$ & $-$ & $-$ & $0$ & $0$ & $0$ & $0$ \\ 
$1$   & $-$ & $-$ & $0$ & $0$ & $0$ & $0$ \\ 
\hline
\end{tabular}
\end{minipage}
\hfill  
\begin{minipage}[!ht]{0.5\textwidth}
\renewcommand{\arraystretch}{1.3}
\centering
\caption{The consistency degrees for the clique $G$ regarding the issue $i_5$ under different values of $\mu$ and $\nu$.}
\label{tab_CM-munu}
\setlength{\tabcolsep}{2.1mm}
\begin{tabular}{ccccccc}
\hline
\diagbox{$\mu$}{$\nu$} & $0$ & $-0.2$ & $-0.4$ & $-0.6$ & $-0.8$ & $-1$ \\ 
\hline
$0$   & $0.64$ & $0.64$ & $0.70$ & $0.70$ & $0.70$ & $0.70$ \\ 
$0.2$ & $0.64$ & $0.64$ & $0.70$ & $0.70$ & $0.70$ & $0.70$ \\ 
$0.4$ & $0.64$ & $0.64$ & $0.70$ & $0.70$ & $0.70$ & $0.70$ \\ 
$0.6$ & $0.64$ & $0.64$ & $0.70$ & $0.70$ & $0.70$ & $0.70$ \\ 
$0.8$ & $0.64$ & $0.64$ & $0.70$ & $0.70$ & $0.70$ & $0.70$ \\ 
$1$   & $0.64$ & $0.64$ & $0.70$ & $0.70$ & $0.70$ & $0.70$ \\ 
\hline
\end{tabular}
\end{minipage}
\end{table}

\begin{table}[!ht]
\centering
\begin{minipage}[!ht]{0.48\textwidth}
\renewcommand{\arraystretch}{1.3}
\centering
\caption{The number of consistency-based feasible strategies $FS^{\mathbb{C}}(G)$ under different values of $\mu$ and $\nu$.}
\label{tab_SC-munu}
\setlength{\tabcolsep}{2.2mm}
\begin{tabular}{ccccccc}
\hline
\diagbox{$\mu$}{$\nu$} & $0$ & $-0.2$ & $-0.4$ & $-0.6$ & $-0.8$ & $-1$ \\ 
\hline
$0$   & $15$ & $15$ & $15$ & $15$ & $15$ & $5$  \\ 
$0.2$ & $19$ & $19$ & $19$ & $19$ & $19$ & $7$ \\  
$0.4$ & $19$ & $19$ & $19$ & $19$ & $19$ & $7$ \\  
$0.6$ & $19$ & $19$ & $19$ & $19$ & $19$ & $7$ \\  
$0.8$ & $19$ & $19$ & $19$ & $19$ & $19$ & $7$ \\  
$1$   & $1$  & $1$  & $1$  & $1$  & $1$  & $0$ \\ 
\hline
\end{tabular}
\end{minipage}
\hfill  
\begin{minipage}[!ht]{0.48\textwidth}
\renewcommand{\arraystretch}{1.3}
\centering
\caption{The number of consistency-based $5$-order feasible strategies $FS_{5}^{\mathbb{C}}(G)$ under different values of $\mu$ and $\nu$.}
\label{tab_SCL-munu}
\setlength{\tabcolsep}{2.2mm}
\begin{tabular}{ccccccc}
\hline
\diagbox{$\mu$}{$\nu$} & $0$ & $-0.2$ & $-0.4$ & $-0.6$ & $-0.8$ & $-1$ \\ 
\hline
$0$   & $0$ & $0$ & $0$ & $0$ & $0$ & $0$ \\ 
$0.2$ & $1$ & $1$ & $1$ & $1$ & $1$ & $0$ \\ 
$0.4$ & $1$ & $1$ & $1$ & $1$ & $1$ & $0$ \\  
$0.6$ & $1$ & $1$ & $1$ & $1$ & $1$ & $0$ \\ 
$0.8$ & $1$ & $1$ & $1$ & $1$ & $1$ & $0$ \\ 
$1$   & $0$ & $0$ & $0$ & $0$ & $0$ & $0$ \\ 
\hline
\end{tabular}
\end{minipage}
\end{table}

\begin{table}[!ht]
\renewcommand{\arraystretch}{1.3}
\begin{center}
\caption{The consistency-based optimal ($5$-optimal) feasible strategies $\widehat{FS}^{\mathbb{C}}(G)$ and $\widehat{FS}_{5}^{\mathbb{C}}(G)$ under different values of $\mu$ and $\nu$.}
\label{tab_opt-munu}
\setlength{\tabcolsep}{0.8mm}
\begin{tabular}{ccccccc}
\hline
\diagbox{$\mu$}{$\nu$} & $0$ & $-0.2$ & $-0.4$ & $-0.6$ & $-0.8$ & $-1$ \\ 
\hline
$0$   & $\emptyset$ & $\emptyset$ & $\emptyset$ & $\emptyset$ & $\emptyset$ & $\emptyset$ \\ 
$0.2$ & $\{\{t_1,t_2,t_3,t_7,t_9\}\}$ & $\{\{t_1,t_2,t_3,t_7,t_9\}\}$ & $\{\{t_1,t_2,t_3,t_7,t_9\}\}$
& $\{\{t_1,t_2,t_3,t_7,t_9\}\}$ & $\{\{t_1,t_2,t_3,t_7,t_9\}\}$ & $\emptyset$ \\  
$0.4$ & $\{\{t_1,t_2,t_3,t_7,t_9\}\}$ & $\{\{t_1,t_2,t_3,t_7,t_9\}\}$ & $\{\{t_1,t_2,t_3,t_7,t_9\}\}$
& $\{\{t_1,t_2,t_3,t_7,t_9\}\}$ & $\{\{t_1,t_2,t_3,t_7,t_9\}\}$ & $\emptyset$ \\   
$0.6$ & $\{\{t_1,t_2,t_3,t_7,t_9\}\}$ & $\{\{t_1,t_2,t_3,t_7,t_9\}\}$ & $\{\{t_1,t_2,t_3,t_7,t_9\}\}$
& $\{\{t_1,t_2,t_3,t_7,t_9\}\}$ & $\{\{t_1,t_2,t_3,t_7,t_9\}\}$ & $\emptyset$ \\    
$0.8$ & $\{\{t_1,t_2,t_3,t_7,t_9\}\}$ & $\{\{t_1,t_2,t_3,t_7,t_9\}\}$ & $\{\{t_1,t_2,t_3,t_7,t_9\}\}$
& $\{\{t_1,t_2,t_3,t_7,t_9\}\}$ & $\{\{t_1,t_2,t_3,t_7,t_9\}\}$ & $\emptyset$ \\      
$1$   & $\emptyset$ & $\emptyset$ & $\emptyset$ & $\emptyset$ & $\emptyset$ & $\emptyset$ \\ 
\hline
\end{tabular}
\end{center}
\end{table}

From Table \ref{tab_R-munu}, as the parameter $\nu$ decreases (indicating either a more stringent condition for judging the overall rating into an opposing attitude or alternatively a more lenient condition for judging the overall rating as a neutral attitude), the overall rating for $G$ regarding the issue $t_5$ undergoes a progressive transformation from oppositional to neutral. The parameter $\mu$ exhibits no discernible influence on the overall rating outcome, demonstrating complete independence from the parameter. 
Since $\rho^{+}_{t_5}(G) = 0.1587$ and $\rho^{-}_{t_5}(G) = 0.4444$, we obtain $\rho^{+}_{t_5}(G) - \rho^{-}_{t_5}(G) < 0$. According to Corollary \ref{corollary3.1}, the overall rating for $G$ on the issue $t_5$ is either neutrality or opposition, with support being mathematically precluded. This quantitative analysis precisely corresponds to the sensitivity analysis results of parameters $\mu$ and $\nu$ in Table \ref{tab_R-munu}. 
From Table \ref{tab_CM-munu}, the consistency degree remains completely unaffected by changes in $\mu$, maintaining a constant value across its entire range from $0$ to $1$. In contrast, the consistency degree increases as $\nu$ decreases from $0$ to $-1$. Notably, the variation laws of the consistency degree and overall rating for the clique $G$ on the issue $t_5$ with the change of parameters $\mu$ and $\nu$ are consistent. Therefore, the overall rating and consistency degree for $G$ regarding the issue $t_5$ are sensitive to the parameter $\nu$ and stable to the parameter $\mu$. 
By Corollary \ref{corollary3.1}, the condition for classifying the overall rating as support depends solely on the parameter $\mu$, whereas the condition for classifying it as opposition depends exclusively on the parameter $\nu$. If the difference between positive and negative power is greater than zero, then the above-mentioned sensitive conclusion will be reversed.

From Tables \ref{tab_SC-munu} and \ref{tab_SCL-munu}, when the parameter $\mu$ reaches its maximum value while $\nu$ simultaneously attains its minimum value (representing the strictest conditions for determining the overall rating as support and opposition), the number of ($5$-order) feasible strategies based on consistency significantly diminishes to zero.
By Tables \ref{tab_SC-munu}-\ref{tab_opt-munu}, when the judgment conditions of the overall ratings are neither the most lenient nor the strictest (i.e. $\mu\neq0,1$ and $\nu\neq0,-1$), the ($5$-order) feasible strategies and the optimal ones remain relatively stable.

\subsubsection{\texorpdfstring{Sensitivity analysis on parameters $\lambda$ and $\tau$}
{Sensitivity analysis on parameters lambda and tau}}\label{5.22jie}\hspace{0.2in}
Considering that the parameter \(\lambda\) pertains exclusively to the consistency-based method, while the parameter \(\tau\) pertains solely to the non-consistency-based method, we conduct a systematic evaluation of the impact of \(\lambda\) on consistency-based feasible strategies and of \(\tau\) on non-consistency-based feasible strategies.

For the clique $G = \{p_1, p_2, p_3, p_6, p_9\}$, assuming $\mu = 0.3$, $\nu = -0.3$, and $L = 5$, 
we examine the influence of the parameter \(\lambda\) on the number of distinct types of consistency-based feasible strategies by varying \(\lambda\) from $0.5$ to $0.95$ in increments of $0.05$. In parallel, to assess the impact of the parameter \(\tau\), we vary its value from $0.05$ to $0.5$ in increments of $0.05$ and analyze the resulting number of distinct types of non-consistency-based feasible strategies. The comprehensive results of this sensitivity analysis for the parameters \(\lambda\) and \(\tau\) are presented in Figures \ref{fig_2} and \ref{fig_3} and Table \ref{tab_opt-munu}.

\begin{figure}[!ht]
  \centering
  \begin{minipage}[c]{0.495\textwidth}
    \centering
    \includegraphics[width=\linewidth]{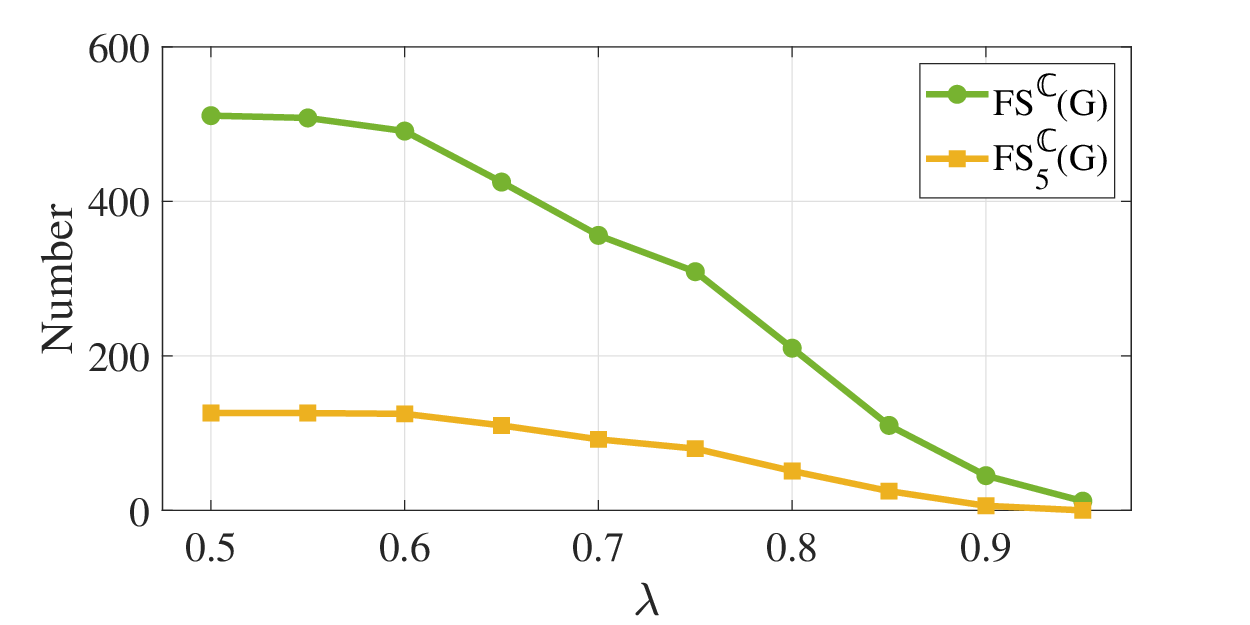}
    \caption{The number of ($5$-optimal) consistency-based feasible strategies under different values of $\lambda$.}
    \label{fig_2}
  \end{minipage}
  \hfill
  \begin{minipage}[c]{0.495\textwidth}
    \centering
    \includegraphics[width=\linewidth]{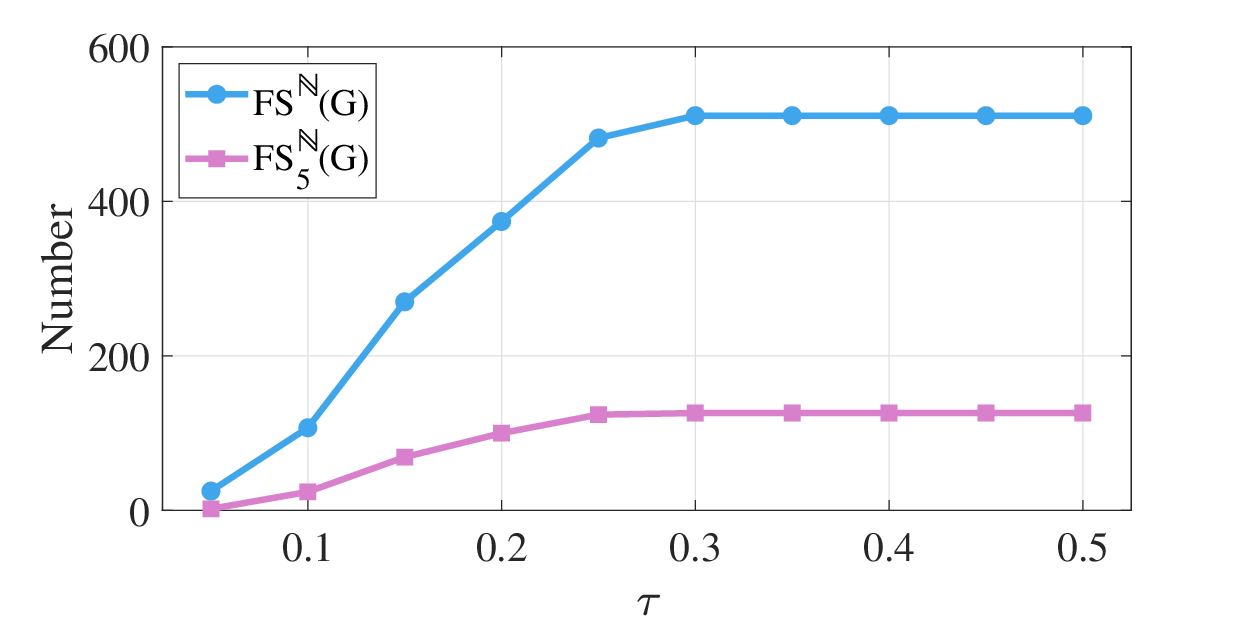}
    \caption{The number of ($5$-optimal) non-consistency-based feasible strategies under different values of $\tau$.}
    \label{fig_3}
  \end{minipage}
\end{figure}

\begin{table}[!ht]
\renewcommand{\arraystretch}{1.3}
\begin{center}
\caption{The consistency-based optimal ($5$-optimal) feasible strategies under different values of $\lambda$ and the non-consistency-based optimal ($5$-optimal) feasible strategies under different values of $\tau$.}
\label{tab_opt-lambdatau}
\setlength{\tabcolsep}{3mm}
\begin{tabular}{ccc|ccc}
\hline
$\lambda$ & $FS^{\mathbb{C}}(G)$ & $FS_{5}^{\mathbb{C}}(G)$ 
& $\tau$ & $FS^{\mathbb{N}}(G)$ & $FS_{5}^{\mathbb{N}}(G)$ \\ 
\hline
$0.50$ & $\{ \{t_1,t_2,t_3,t_7,t_9\} \}$ & $\{ \{t_1,t_2,t_3,t_7,t_9\} \}$ 
& $0.05$ & $\{ \{t_1,t_2,t_3,t_7,t_9\} \}$ & $\{ \{t_1,t_2,t_3,t_7,t_9\} \}$ \\  
$0.55$ & $\{ \{t_1,t_2,t_3,t_7,t_9\} \}$ & $\{ \{t_1,t_2,t_3,t_7,t_9\} \}$ 
& $0.10$ & $\{ \{t_1,t_2,t_3,t_7,t_9\} \}$ & $\{ \{t_1,t_2,t_3,t_7,t_9\} \}$ \\  
$0.60$ & $\{ \{t_1,t_2,t_3,t_7,t_9\} \}$ & $\{ \{t_1,t_2,t_3,t_7,t_9\} \}$ 
& $0.15$ & $\{ \{t_1,t_2,t_3,t_7,t_9\} \}$ & $\{ \{t_1,t_2,t_3,t_7,t_9\} \}$ \\  
$0.65$ & $\{ \{t_1,t_2,t_3,t_7,t_9\} \}$ & $\{ \{t_1,t_2,t_3,t_7,t_9\} \}$ 
& $0.20$ & $\{ \{t_1,t_2,t_3,t_7,t_9\} \}$ & $\{ \{t_1,t_2,t_3,t_7,t_9\} \}$ \\    
$0.70$ & $\{ \{t_1,t_2,t_3,t_7,t_9\} \}$ & $\{ \{t_1,t_2,t_3,t_7,t_9\} \}$ 
& $0.25$ & $\{ \{t_1,t_2,t_3,t_7,t_9\} \}$ & $\{ \{t_1,t_2,t_3,t_7,t_9\} \}$ \\    
$0.75$ & $\{ \{t_1,t_2,t_3,t_7,t_9\} \}$ & $\{ \{t_1,t_2,t_3,t_7,t_9\} \}$
& $0.30$ & $\{ \{t_1,t_2,t_3,t_7,t_9\} \}$ & $\{ \{t_1,t_2,t_3,t_7,t_9\} \}$ \\ 
$0.80$ & $\{ \{t_1,t_2,t_3,t_7,t_9\} \}$ & $\{ \{t_1,t_2,t_3,t_7,t_9\} \}$ 
& $0.35$ & $\{ \{t_1,t_2,t_3,t_7,t_9\} \}$ & $\{ \{t_1,t_2,t_3,t_7,t_9\} \}$ \\
$0.85$ & $\{ \{t_1,t_2,t_3,t_7,t_9\} \}$ & $\{ \{t_1,t_2,t_3,t_7,t_9\} \}$ 
& $0.40$ & $\{ \{t_1,t_2,t_3,t_7,t_9\} \}$ & $\{ \{t_1,t_2,t_3,t_7,t_9\} \}$ \\
$0.90$ & $\{ \{t_1,t_2,t_3,t_7,t_9\} \}$ & $\{ \{t_1,t_2,t_3,t_7,t_9\} \}$ 
& $0.45$ & $\{ \{t_1,t_2,t_3,t_7,t_9\} \}$ & $\{ \{t_1,t_2,t_3,t_7,t_9\} \}$ \\
$0.95$ & $\emptyset$ & $\emptyset$ 
& $0.50$ & $\{ \{t_1,t_2,t_3,t_7,t_9\} \}$ & $\{ \{t_1,t_2,t_3,t_7,t_9\} \}$ \\
\hline
\end{tabular}
\end{center}
\end{table}

As illustrated in Figures \ref{fig_2} and \ref{fig_3}, an increase in the parameter \(\lambda\) results in a decrease in the number of consistency-based (5-order) feasible strategies. In contrast, the number of non-consistency-based (5-order) feasible strategies increases with higher values of \(\tau\). Notably, when \(\tau\) exceeds $0.3$, the number of non-consistency-based feasible strategies stabilizes and no longer changes.
According to Table \ref{tab_opt-lambdatau}, the (5-order) optimal feasible strategy remains stable and robust across different values of \(\lambda\) and \(\tau\). Furthermore, the consistency-based optimal feasible strategy coincides with its non-consistency-based counterpart under all parameter settings, except when \(\lambda = 0.95\), where neither optimal nor 5-order feasible strategies can be identified.
In summary, consistency-based feasible strategies exhibit sensitivity to variations in \(\lambda\), while non-consistency-based feasible strategies are influenced by changes in \(\tau\). Nevertheless, both types of optimal feasible strategies demonstrate considerable robustness with respect to these parameters.
}

\subsection{Comparison analysis}\label{5.3jie}\hspace{0.2in}
In this section, we implement some comparison analysis to demonstrate the effectiveness and superiority of our proposed models.

\subsubsection{Comparisons to Xu's conflict analysis model}\label{5.31jie}\hspace{0.2in}
We compare our proposed model with consistency measures to the model presented by Xu~\cite{Xu2022} using the case study of NBA labor negotiations. Additionally, we analyze the impact of various weights and parameters on overall ratings and feasible strategies.

To be consistent, we set the following parameters in Xu's model: 
$L = 3$, $\gamma_P = 0.5$, and $\lambda = 0.966$. 
In our proposed model, the parameters are specified as:
$\Theta = (0.25, 0.18, 0.07, 0.06, 0.05, 0.1, 0.09, 0.12, 0.03, 0.05)$,
$W = (0.3, 0.03, 0.02, 0.2, 0.15, 0.1, 0.1, 0.05, 0.05)$,
$\mu = 0.4, \, \nu = -0.5, \, \gamma_P = 0.5, \, \lambda = 0.966.$
For the subsequent analysis, we select three cliques: $\{p_1, p_2, p_3, p_6, p_9\}$, $\{p_2, p_4, p_6, p_7, p_9\}$, and $\{p_4, p_5, p_7, p_9, p_{10}\}$. For each clique, we compute the overall ratings, feasible strategies, and corresponding consistency degrees.
The overall ratings for Xu's model are presented in Table~\ref{tab_R-Xu}, while those for our proposed model are displayed in Table~\ref{tab_R-Our}. In Tables \ref{tab_R-Xu} and \ref{tab_R-Our}, ``$+1$'' and ``$-1$'' are omitted as ``$+$'' and ``$-$'', respectively. The 3-order dominant feasible strategies and their corresponding consistency degrees for Xu's model are provided in Table~\ref{tab_FS-Xu}. In comparison, the feasible strategies and their corresponding consistency degrees for our proposed model are shown in Table~\ref{tab_FS-Our}.

\begin{table}[!ht]
\renewcommand{\arraystretch}{1.3}
\begin{center}
\caption{The overall ratings by Xu's conflict analysis model in NBA labor negotiations.}
\label{tab_R-Xu}
\setlength{\tabcolsep}{5.3mm}
\begin{tabular}{cccccccccc}
\hline
$G$ & $t_{1}$ & $t_{2}$ & $t_{3}$ & $t_{4}$ & $t_{5}$ & $t_{6}$ & $t_{7}$ & $t_{8}$ & $t_{9}$\\
\hline
$\{p_{1},p_{2},p_{3},p_{6},p_{9}\}$  
&$+$ & $+$ & $0$ & {\cellcolor{blue!20}$+$} & $0$ & $0$ & $-$ & \cellcolor{blue!20}$-$ & \cellcolor{blue!20}$-$ \\
$\{p_{2},p_{4},p_{6},p_{7},p_{9}\}$  
&$+$ & $+$ & $0$ & $+$ & \cellcolor{blue!20}$-$ & \cellcolor{blue!20}$+$ & \cellcolor{blue!20}$-$ & $+$ & $-$ \\
$\{p_{4},p_{5},p_{7},p_{9},p_{10}\}$ 
&\cellcolor{blue!20}$+$ & $+$ & $+$ & \cellcolor{blue!20}$+$ & $+$ & $0$ & $+$ & \cellcolor{blue!20}$-$ & $-$ \\
\hline
\end{tabular}
\end{center}
\end{table}

\begin{table}[!ht]
\renewcommand{\arraystretch}{1.3}
\begin{center}
\caption{The overall ratings by the proposed model with consistency measures in NBA labor negotiations.}
\label{tab_R-Our}
\setlength{\tabcolsep}{5.3mm}
\begin{tabular}{cccccccccc}
\hline
$G$ & $t_{1}$ & $t_{2}$ & $t_{3}$ & $t_{4}$ & $t_{5}$ & $t_{6}$ & $t_{7}$ & $t_{8}$ & $t_{9}$\\
\hline
$\{p_{1},p_{2},p_{3},p_{6},p_{9}\}$  
&$+$ & $+$ & $0$ & \cellcolor{blue!20}$0$ & $0$ & $0$ & $-$ & \cellcolor{blue!20}$0$ & \cellcolor{blue!20}$0$ \\
$\{p_{2},p_{4},p_{6},p_{7},p_{9}\}$  
&$+$ & $+$ & $0$ & $+$ & \cellcolor{blue!20}$0$ & \cellcolor{blue!20}$0$ & \cellcolor{blue!20}$0$ & $+$ & $-$ \\
$\{p_{4},p_{5},p_{7},p_{9},p_{10}\}$ 
&\cellcolor{blue!20}$0$ & $+$ & $+$ & \cellcolor{blue!20}$0$ & $+$ & $0$ & $+$ & \cellcolor{blue!20}$0$ & $-$ \\
\hline
\end{tabular}
\end{center}
\end{table}

\begin{table}[!ht]
\renewcommand{\arraystretch}{1.3}
\begin{center}
\caption{The 3-order dominant feasible strategies and consistency degrees by Xu's model in NBA labor negotiations.}\label{tab_FS-Xu}
\setlength{\tabcolsep}{2mm}
\begin{tabular}{cccccccccccc}
\hline
$G$ & $FS_{3}^{Xu}(G)$ & $\mathbb{CM}^{Xu}_{FS_{3}^{Xu}(G)}(G)$
& \multicolumn{9}{c}{$\mathbb{CM}^{Xu}_{t}(G)~(t=t_1,t_2,\dots,t_9)$}\\
\hline
$\{p_{1},p_{2},p_{3},p_{6},p_{9}\}$  & $\{t_{1},t_{2},t_{7}\}$ & $0.966$ 
&$\bf{1}$ & $\bf{0.9}$ & $0.6$ & $0.7$ & $0.6$ & $0.6$ & $\bf{1}$ & $0.6$ & $0.6$ \\
$\{p_{2},p_{4},p_{6},p_{7},p_{9}\}$  & $\{t_{1},t_{2},t_{9}\}$ & $0.966$ 
&$\bf{1}$ & $\bf{1}$ & $0.6$ & $0.7$ & $0.6$ & $0.7$ & $0.6$ & $0.6$ & $\bf{0.9}$ \\
$\{p_{4},p_{5},p_{7},p_{9},p_{10}\}$ & $\{t_{2},t_{3},t_{9}\}$ & $0.966$ 
&$0.6$ & $\bf{1}$ & $\bf{0.9}$ & $0.6$ & $0.8$ & $0.6$ & $0.6$ & $0.8$ & $\bf{1}$ \\
\hline
\end{tabular}
\end{center}
\end{table}

\begin{table}[!ht]
\renewcommand{\arraystretch}{1.3}
\begin{center}
\caption{The feasible strategies and consistency degrees by the proposed model with consistency measures in NBA labor negotiations.}\label{tab_FS-Our}
\setlength{\tabcolsep}{9.5mm}
\begin{tabular}{cclc}
\hline
$G$ & $\#\big(FS^{\mathbb{C}}(G)\big)$ & $J\in FS^{\mathbb{C}}(G)$ & $\mathbb{CM}(G,J)$ \\
\hline
$\{p_{1},p_{2},p_{3},p_{6},p_{9}\}$  & 10
&\makecell[l]{$\{t_{1}\}$ \\ $\{t_{7}\}$ \\ 
$\{t_{1},t_{2}\}$ \\ $\{t_{1},t_{3}\}$ \\ $\{t_{1},t_{7}\}$ \\ $\{t_{2},t_{7}\}$ \\ 
$\{t_{1},t_{2},t_{3}\}$ \\ $\{t_{1},t_{2},t_{7}\}$ \\ $\{t_{1},t_{3},t_{7}\}$ \\ $\{t_{1},t_{2},t_{3},t_{7}\}$} 
&\makecell[c]{$1$ \\ $1$ \\ $0.9949$ \\ $0.9702$ \\ $1$ \\ $0.9872$ \\ $0.9680$ \\ $0.9961$ \\ $0.9773$ \\ $0.9751$} \\
\hline
$\{p_{2},p_{4},p_{6},p_{7},p_{9}\}$  & 9
&\makecell[l]{$\{t_{1}\}$ \\ $\{t_{2}\}$ \\ 
$\{t_{1},t_{2}\}$ \\ $\{t_{1},t_{3}\}$ \\ $\{t_{1},t_{8}\}$ \\ $\{t_{1},t_{9}\}$ \\ 
$\{t_{1},t_{2},t_{3}\}$ \\ $\{t_{1},t_{2},t_{8}\}$ \\ $\{t_{1},t_{2},t_{9}\}$}
&\makecell[c]{$1$ \\ $1$ \\ $1$ \\ $0.9708$ \\ $0.9674$ \\ $0.9845$ \\ $0.9733$\\ $0.97$\\ $0.9857$} \\
\hline
$\{p_{4},p_{5},p_{7},p_{9},p_{10}\}$ & 6
&\makecell[l]{$\{t_{2}\}$ \\ $\{t_{9}\}$ \\ 
$\{t_{2},t_{3}\}$ \\ $\{t_{2},t_{9}\}$ \\ $\{t_{3},t_{9}\}$ \\
$\{t_{2},t_{3},t_{9}\}$}
&\makecell[c]{$1$ \\ $1$ \\ $0.9786$ \\ $1$ \\ $0.9847$ \\ $0.9893$} \\
\hline
\end{tabular}
\end{center}
\end{table}

As our approach considers the significance of the difference between positive and negative degrees, some positive and negative overall ratings in Xu's model are deemed neutral in ours, as highlighted in Tables \ref{tab_R-Xu} and \ref{tab_R-Our}. The incorporation of two thresholds has rendered the conditions for classifying overall ratings as positive or negative more stringent, while simultaneously relaxing the criteria for categorizing them as neutral. Consequently, the appropriate adjustment of these thresholds can adapt the error tolerance of our model.

In Table~\ref{tab_FS-Xu}, for Xu's model, the 3-order dominant feasible strategies consist of subsets of issues whose consistency degrees are among the top three across all issues. Notably, the consistency degrees of all 3-order dominant feasible strategies are $0.966$. 
From Table \ref{tab_FS-Our}, by considering only the $3$-order feasible strategies, we observe that their consistency degrees are all greater than $0.966$. This demonstrates that, by incorporating agent weights, issue weights, and thresholds for evaluating overall ratings, our model yields feasible strategies with higher consistency degrees compared to Xu's model.
For the clique $\{p_{1},p_{2},p_{3},p_{6},p_{9}\}$, if a decision-maker wishes to choose three issues as feasible strategies, Xu's model offers only one option, whereas our proposed model provides three options. Additionally, if the decision-maker aims to select one, two, or four issues as feasible strategies, Xu's model does not provide any solutions, while our model offers multiple solutions. The same advantage holds for the cliques $\{p_{2},p_{4},p_{6},p_{7},p_{9}\}$ and $\{p_{4},p_{5},p_{7},p_{9},p_{10}\}$.

\subsubsection{Comparisons to ten conflict analysis models}\label{5.32jie}\hspace{0.2in}
We perform a comparative analysis of our presented models and ten existing models, including Pawlak's model~\cite{Pawlak1998}, Yao's model~\cite{Yao2019}, {\color{black}Zhi's model~\cite{Zhi2020}}, Sun's model~\cite{Sun2016}, Sun's model~\cite{Sun2020}, {\color{black}Lang's model~\cite{Lang2020-KBS}}, Xu's model~\cite{Xu2022}, Du's model~\cite{Du2022},  Yang's model~\cite{Yang2023}, and Li's model~\cite{Li2023}. The comparison of these models is summarized in Table~\ref{tab_compare}, where ``$\checkmark$'' indicates that the model incorporates the corresponding aspect, while the ``$\times$'' denotes otherwise.

\begin{table}[!ht]
\renewcommand{\arraystretch}{1.3}
\begin{center}
\caption{Comparison of our models and ten conflict analysis models.}
\label{tab_compare}
\setlength{\tabcolsep}{1mm}
\scalebox{0.92}{
\begin{tabular}{llcclc}
\hline
Models & Main measures & \makecell[c]{Agent\\weights} & \makecell[c]{Issue\\weights} & \makecell[l]{Construction of \\ feasible strategies} & \makecell[c]{Optimal feasible \\ strategy} \\ 
\hline
\makecell[l]{~Our models} & \makecell[l]{Consistency and \\non-consistency measures} & $\checkmark$ & $\checkmark$ & \makecell[l]{Feasible strategies and \\ $L$-order feasible strategies} & $\checkmark$ \\
\rule{0pt}{18pt}
Pawlak's model \cite{Pawlak1998} & Distance function & $\times$ & $\times$  & \makecell[c]{$\times$} & $\times$ \\
\rule{0pt}{18pt}
Yao's model \cite{Yao2019} & Distance function & $\times$ & $\times$  & \makecell[c]{$\times$} & $\times$ \\
\rule{0pt}{24pt}
Zhi's model \cite{Zhi2020} & \makecell[l]{Consistency and \\inconsistency measures} & $\times$ & $\times$  & \makecell[c]{$\times$} & $\times$ \\
\rule{0pt}{18pt}
Sun's model \cite{Sun2016} & Two set-valued mappings  & $\times$ & $\times$  & Feasible strategies & $\checkmark$ \\
\rule{0pt}{24pt}
Sun's model \cite{Sun2020} & \makecell[l]{Probabilistic lower and\\upper approximations} & $\times$ & $\times$ & \makecell[l]{Feasible consensus strategies} & $\checkmark$ \\
\rule{0pt}{24pt}
Lang's model \cite{Lang2020-KBS} & \makecell[l]{Conditional support and\\opposition evaluations} & $\times$ & $\times$ & \makecell[l]{Possible strategies} & $\checkmark$ \\
\rule{0pt}{18pt}
Xu's model \cite{Xu2022} & Consistency measures & $\times$ & $\times$ & \makecell[l]{$L$-order dominant feasible strategies} & $\times$ \\
\rule{0pt}{18pt}
Du's model \cite{Du2022} & Conflict distance & $\times$ & $\checkmark$ & Feasible strategies & $\checkmark$ \\
\rule{0pt}{18pt}
Yang's model \cite{Yang2023} & Conflict distance & $\times$ & $\checkmark$ & Feasible strategies & $\checkmark$ \\
\rule{0pt}{18pt}
Li's model \cite{Li2023} & Conflict distance & $\times$ & $\checkmark$ & Feasible strategies & $\checkmark$ \\
\hline
\end{tabular}}
\end{center}
\end{table}

Pawlak's model~\cite{Pawlak1998} and Yao's model~\cite{Yao2019} analyze conflict situations using distance functions. {\color{black}Zhi's model~\cite{Zhi2020} utilizes consistency and inconsistency measures to investigate conflict analysis under one-vote veto. However, none of these three models provides explicit solutions for resolving conflicts.}
Sun's models~\cite{Sun2016, Sun2020} perform conflict analysis through rough sets and probabilistic rough sets over two universes, respectively. {\color{black}Lang's model \cite{Lang2020-KBS} establishes qualitative three-way conflict analysis models based on conditional support and opposition evaluations. However, these three models share a common limitation in their disregard for the weights of agents and issues.}
Xu's model~\cite{Xu2022} introduces two consistency measures for a clique concerning a single issue and an issue set and accordingly selects $L$-order dominant feasible strategies. However, Xu's consistency measures neglect the weights of agents and issues. Moreover, the model does not provide a mechanism for selecting an optimal feasible strategy from multiple $L$-order dominant feasible strategies.
Du's model~\cite{Du2022}, Yang's model~\cite{Yang2023}, and Li's model~\cite{Li2023} consider feasible strategies to be any subsets of the union of a weak-conflict and a non-conflict issue set. They select the optimal one with the maximum value from all evaluation function values. These models focus solely on issue weights and do not account for agent weights.
{\color{black}In contrast, our two proposed models incorporate both agent and issue weights, and identify the ($L$-order) feasible strategies and the ($L$-order) optimal ones from two distinct perspectives of consistency and non-consistency. As a result, our model demonstrates more comprehensive performance and superior capabilities compared to these ten baseline models.}

\section{Conclusions}\label{6jie}\hspace{0.2in}
In this paper, we explored feasible strategies in three-way conflict analysis for three-valued situation tables, focusing on both consistency and non-consistency perspectives. We defined an overall rating function and developed two consistency measures that incorporate both agent and issue weights. These measures extended the consistency measures proposed by Xu~\cite{Xu2022}, and we demonstrated methods for deriving feasible strategies, $L$-order feasible strategies, the optimal feasible strategy, and $L$-order optimal feasible strategies for a clique using the consistency measures. Additionally, we introduced conflict measures for both single and multiple issues, considering the weights of agents and issues, and used them to analyze the trisection of agent pairs. We also derived two non-consistency measures that take into account agent and issue weights and developed corresponding feasible strategies, $L$-order feasible strategies, the optimal feasible strategy, and $L$-order optimal feasible strategies for a clique using the non-consistency measures.
Algorithms were designed for identifying ($L$-order) feasible strategies and ($L$-order) optimal feasible strategies using both consistency and non-consistency measures. These models were applied to NBA labor negotiations and Gansu Province's development plans, demonstrating their practical applicability. A sensitivity analysis was conducted to reveal the influence of each parameter on the selection of feasible strategies and a comparative analysis with eight existing conflict analysis models was also executed to validate the superiority of our presented models.

In {\color{black}{future work}}, we intend to extend this study to more complicated situation tables, including many-valued, fuzzy, multi-scale, hybrid, and incomplete situation tables, to broaden the applicability of the proposed models. {\color{black}We will further investigate  conflict resolution based on both consistency and non-consistency measures simultaneously.} We also aim to develop more efficient algorithms for selecting feasible strategies to resolve and mitigate conflicts. 
Another further potential avenue for {\color{black}{future research}} is the integration of conflict analysis with game theory, which has the potential to significantly enhance both the theoretical foundations and practical applications of our approach.

\section*{Acknowledgements}\hspace{0.2in}
This work is supported by the National Natural Science Foundation of China (Nos. 62076040, 12471431, 62276217), the Scientific Research Fund of Hunan Provincial Education Department (No. 22A0233), and a Discovery Grant from the Natural Sciences and Engineering Research Council of Canada.

\section*{Appendix A. Proof of Theorem \ref{theorem3.1}}\hspace{0.2in}
\vspace{-7mm}
\begin{proof}
	For a clique $G \subseteq P$ and an issue $t \in T$, we have the following three cases: 
	\begin{itemize}[labelindent = 0em, leftmargin = *]
		\setlength{\itemsep}{0pt}
		\item \textbf{Case 1: $R(G,t) = +1$}. We have: 
		\begin{align*}
			\mathbb{CM}(G,t) 
			&= 1 - \frac{1}{2} \sum_{p \in G} \big( \theta(p|G) \times |r(p,t) - (+1)| \big) \\
			&= 1 - \frac{1}{2} \left[ 
			\sum_{p \in G_{t}^{+}} \big( \theta(p|G) \times |1 - 1| \big) 
			+ \sum_{p \in G_{t}^{0}} \big( \theta(p|G) \times |0 - 1| \big) 
			+ \sum_{p \in G_{t}^{-}} \big( \theta(p|G) \times |-1 - 1| \big)
			\right] 
            \displaybreak[1] \\
			&= 1 - \frac{1}{2} \left( 0 + \sum_{p \in G_{t}^{0}} \theta(p|G) + 2 \sum_{p \in G_{t}^{-}} \theta(p|G) \right) \displaybreak[1] \\
			&= 1 - \frac{\rho_{t}^{0}(G) + 2 \rho_{t}^{-}(G)}{2} \\
			&= \rho_{t}^{+}(G) + \frac{1}{2} \rho_{t}^{0}(G).
		\end{align*}
        
		\item \textbf{Case 2: $R(G,t) = 0$}. We have: 
		\begin{align*}
			\mathbb{CM}(G,t) 
			&= 1 - \frac{1}{2} \sum_{p \in G} \big( \theta(p|G) \times |r(p,t) - 0| \big) 
			\displaybreak[1]\\
			&= 1 - \frac{1}{2} \left[ 
			\sum_{p \in G_{t}^{+}} \big( \theta(p|G) \times |1 - 0| \big)
			+ \sum_{p \in G_{t}^{0}} \big( \theta(p|G) \times |0 - 0| \big)
			+ \sum_{p \in G_{t}^{-}} \big( \theta(p|G) \times |-1 - 0| \big)
			\right] \\
			&= 1 - \frac{1}{2} \left( \sum_{p \in G_{t}^{+}} \theta(p|G) + 0 + \sum_{p \in G_{t}^{-}} \theta(p|G) \right) \displaybreak[1]\\
			&= 1 - \frac{\rho_{t}^{+}(G) + \rho_{t}^{-}(G)}{2} \\
			&= \frac{1}{2} \rho_{t}^{+}(G) + \rho_{t}^{0}(G) + \frac{1}{2} \rho_{t}^{-}(G).
		\end{align*}
        
		\item \textbf{Case 3: $R(G,t) = -1$}. We have:
		\begin{align*}
			\mathbb{CM}(G,t) 
			&= 1 - \frac{1}{2} \sum_{p \in G} \big( \theta(p|G) \times |r(p,t) - (-1)| \big) \\
			&= 1 - \frac{1}{2} \left[ 
			\sum_{p \in G_{t}^{+}} \big( \theta(p|G) \times |1 + 1| \big)
			+ \sum_{p \in G_{t}^{0}} \big( \theta(p|G) \times |0 + 1| \big)
			+ \sum_{p \in G_{t}^{-}} \big( \theta(p|G) \times |-1 + 1| \big)
			\right]\
            \displaybreak[1]\\
			&= 1 - \frac{1}{2} \left( 2 \sum_{p \in G_{t}^{+}} \theta(p|G) + \sum_{p \in G_{t}^{0}} \theta(p|G) + 0 \right) \\
			&= 1 - \frac{2 \rho_{t}^{+}(G) + \rho_{t}^{0}(G)}{2} \\
			&= \rho_{t}^{-}(G) + \frac{1}{2} \rho_{t}^{0}(G).
		\end{align*}
	\end{itemize}
\end{proof}

\section*{Appendix B. Proof of Theorem \ref{theorem3.2}}\hspace{0.2in}
\vspace{-7mm}
\begin{proof}
	We prove that \( \mathbb{CM}(G,t) \geq 0.5 \) by considering the following cases.
	\begin{itemize}[labelindent = 0em, leftmargin = *]
		\setlength{\itemsep}{0pt}
		\item \textbf{Case 1: $R(G,t) = +1$}. By Corollary \ref{corollary3.1}, we have $\rho_{t}^{+}(G) - \rho_{t}^{-}(G) > \mu \geq 0$, which implies that $\rho_{t}^{+}(G) > \rho_{t}^{-}(G)$. Therefore, we can conclude: 
        \begin{align*}
        \mathbb{CM}(G,t) = \frac{2 \rho_{t}^{+}(G) + \rho_{t}^{0}(G)}{2} > \frac{\rho_{t}^{+}(G) + \rho_{t}^{0}(G) + \rho_{t}^{-}(G)}{2} = 0.5.
        \end{align*}
        
		\item \textbf{Case 2: $R(G,t) = 0$}. By Proposition \ref{proposition_rho}, we know that $\rho_{t}^{0}(G) \geq 0$. Hence, we have:
        \begin{align*}
        \mathbb{CM}(G,t) = \frac{\rho_{t}^{+}(G) + 2 \rho_{t}^{0}(G) + \rho_{t}^{-}(G)}{2} \geq \frac{\rho_{t}^{+}(G) + \rho_{t}^{0}(G) + \rho_{t}^{-}(G)}{2} = 0.5.    
        \end{align*}
        
		\item \textbf{Case 3: $R(G,t) = -1$}. By Corollary \ref{corollary3.1}, we have $\rho_{t}^{+}(G) - \rho_{t}^{-}(G) < \nu \leq 0$, which implies that $\rho_{t}^{+}(G) < \rho_{t}^{-}(G)$. Therefore, we conclude:
        \begin{align*}
        \mathbb{CM}(G,t) = \frac{2 \rho_{t}^{-}(G) + \rho_{t}^{0}(G)}{2} > \frac{\rho_{t}^{+}(G) + \rho_{t}^{0}(G) + \rho_{t}^{-}(G)}{2} = 0.5.
        \end{align*}
	\end{itemize}
	In all cases, we have shown that $\mathbb{CM}(G,t) \geq 0.5$, so $\mathbb{CM}(G,t) \in [0.5, 1]$.
	Furthermore, since $\omega(t|J) \in [0, 1]$, it follows that:
    \begin{align*}
    \mathbb{CM}(G,J) = \sum_{t \in J} \big( \omega(t|J) \times \mathbb{CM}(G,t) \big) \in [0.5, 1].    
    \end{align*}
\end{proof}

\bibliographystyle{elsarticle-num}
\bibliography{references}

\end{document}